%% file: iclr2027_conference.tex
\documentclass{article} 
\usepackage{iclr2027_conference,times}

\input{math_commands.tex}

\usepackage{hyperref}
\usepackage{url}

\usepackage{algpseudocode}
\usepackage{amsmath}
\usepackage{multirow}
\usepackage{pifont}
\usepackage{tcolorbox}
\usepackage{verbatim}
\tcbuselibrary{breakable}
\usepackage{booktabs} 
\usepackage{colortbl}
\usepackage{multicol}
\usepackage{makecell}
\usepackage{tabularx}
\definecolor{bgmark}{gray}{0.9} 
\usepackage{enumitem}

\usepackage{xspace}  
\newcommand{\sys}{\texttt{CauTion}\xspace}
\definecolor{morandiorange}{HTML}{FDE6D0}
  \usepackage{threeparttable}

\newcommand{\claude}{Claude-Sonnet-4.6\xspace}

\newenvironment{pseudocode}[2]{
  \def\pseudocodecaption{#1}
  \def\pseudocodelabel{#2}
  \begin{figure}[htbp]
  \centering
  \setlength{\tabcolsep}{10pt}
  \begin{tabular}{|c|}
  \hline
  \begin{minipage}{0.95\textwidth}
  \vspace{6pt}
}{
  \vspace{6pt}
  \end{minipage} \\
  \hline
  \end{tabular}
  \caption{\pseudocodecaption}
  \label{\pseudocodelabel}
  \end{figure}
}

\title{\textsc{CauTion}: Knowing When to Trust LLMs for Ensemble Causal Discovery}

\author{Bo Peng$^{1,2,3,}$\thanks{Equal contribution.}\hspace{0.3em}, Kaiwen Wu$^{1,4,*}$, Sirui Chen$^{1,5}$,
\textbf{Zhiheng Wang$^{1,3}$, Yu Qiao$^{1,2}$, Chaochao Lu$^{1,}$\thanks{Corresponding author.}} \\
$^{1}$Shanghai AI Laboratory, $^{2}$Shanghai Innovation Institute, 
$^{3}$Shanghai Jiao Tong University, \\$^{4}$Nanjing University, 
$^{5}$Tongji University
}

\iclrfinalcopy 
\begin{document}

\maketitle
\fancyhead{}
\renewcommand{\headrulewidth}{0pt}

\begin{abstract}
Causal discovery from observational data remains challenging due to the fundamental limitations of purely statistical methods, such as statistical distinguishability within equivalence classes and sensitivity to finite sample sizes. While large language models (LLMs) offer a promising source of domain knowledge to complement statistical inference, existing LLM-augmented methods are vulnerable to LLM errors. A central challenge is therefore deciding when to trust LLM knowledge and when to trust statistical evidence. We propose \sys, a trust-calibrated ensemble framework that addresses this challenge through annotation-free reliability estimation and selective arbitration. \sys proceeds in three stages. First, an algorithm ensemble utilizes a consensus voting to resolve up to 96\% of edges on which algorithms agree, achieving near-perfect accuracy on the filtered consensus edges. Second, a trust-calibrated arbitration mechanism estimates the relative reliability of the LLM and the algorithms via an annotation-free trust calibration procedure, which is then utilized to govern a trust-weighted voting process that restricts LLM arbitration exclusively to edges with unreliable algorithmic evidence. Third, a cycle repair step is applied to guarantee the final causal graph is validly acyclic. Experiments on six datasets demonstrate that \sys consistently outperforms both data-centric and LLM-augmented baselines in SHD and F1, with strong robustness to LLM errors.
\end{abstract}

\begin{figure}[htbp]
  \centering
  \includegraphics[width=1\textwidth]{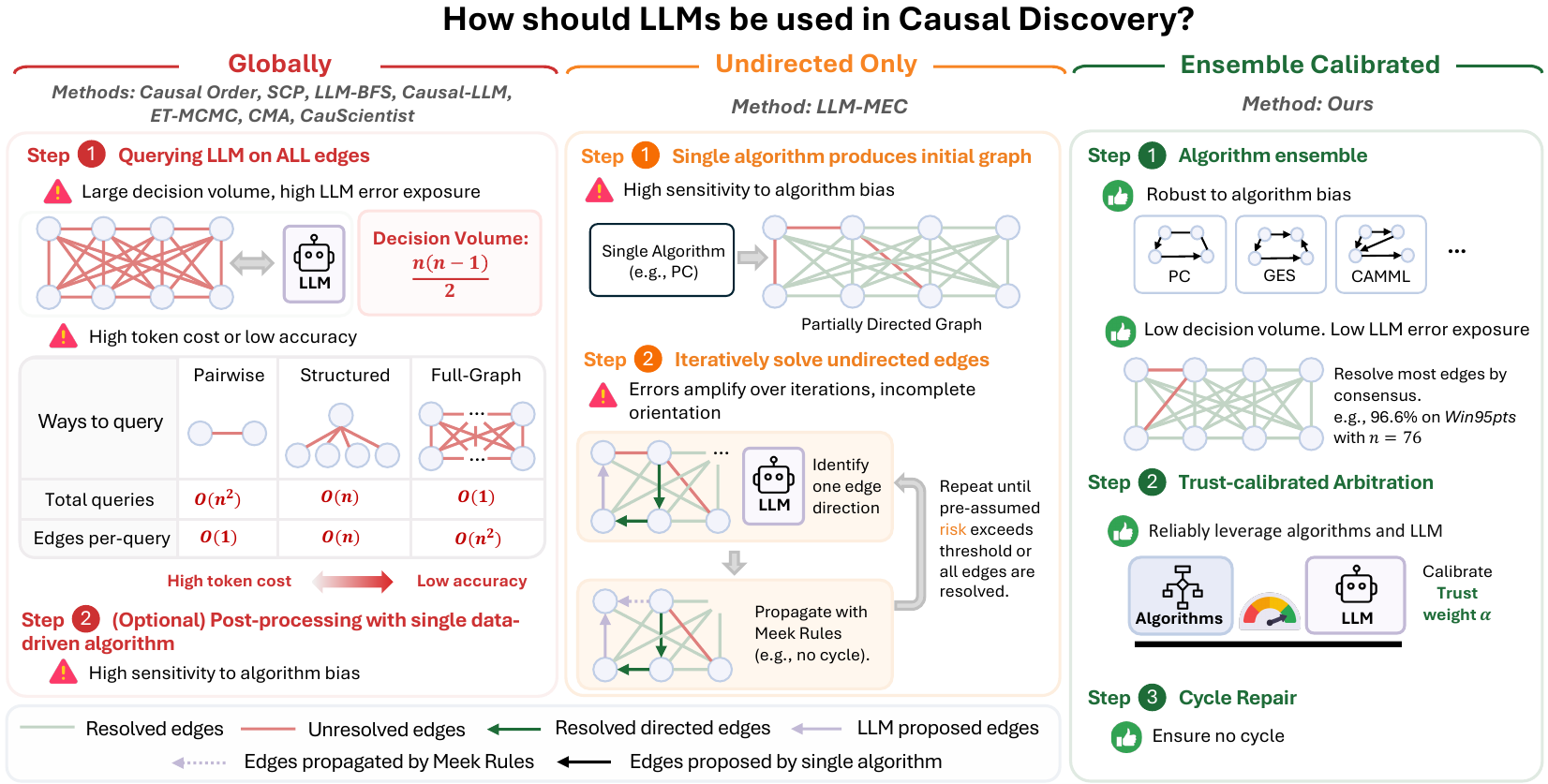}
  \caption{\textbf{The challenge of deciding when to trust LLM knowledge
and data-centric algorithms in causal discovery.} \textbf{Global methods} expose $O(n^2)$ pairwise decisions to potentially unreliable LLM judgments, and incurs either high token cost or low accuracy due to high reasoning complexity per-query.
\textbf{Undirected-only methods} restrict LLM intervention to unresolved orientations, but suffer from error amplification over iterations and incomplete orientation.
In both settings, reliance on a single data-centric algorithm can further expose the final graph to algorithm-specific errors. On the other hand, \sys addresses these limitations via algorithm ensemble and trust-calibrated arbitration, enabling reliable integration robust to both LLM and single algorithm errors.
  }
  \label{fig:intro}
\end{figure}

\section{Introduction}
Causal discovery serves as a cornerstone for scientific inquiry and robust 
artificial intelligence \citep{spirtes2000causation, pearl2009causality, pearl2018book, yao2021survey}. 
Existing methods, including constraint-based approaches 
\citep{spirtes2000causation, 10.5555/2074158.2074215}, score-based 
methods \citep{chickering2002optimal, andrews2023fast, lam2022grasp}, 
and continuous optimization \citep{zheng2018dags, bello2022dagma}, have made substantial progress on causal recovery. However, they remain fundamentally limited: constraint-based methods rely on sufficient sample sizes and can only recover a Markov equivalence class (MEC)~\citep{verma1990equivalence}; score-based and continuous optimization methods face the local minimum problem and are sensitive to the choice of score function under specific assumptions and hyperparameter choices.

Trained on vast corpora, LLMs encode rich prior knowledge about real-world mechanisms and can reason about causal relationships from variable semantics.
LLM-augmented methods \citep{jiralerspong2024efficient,roy2025causal,10966043,abdulaal2023causal,peng2026causcientist,vashishtha2025causal,takayama2025integrating,long2023causal} therefore use such semantic priors as an additional source of evidence to compensate for the limitations of purely data-centric causal discovery algorithms. 
Existing LLM-augmented causal discovery methods can be broadly grouped into two categories: (1) \textbf{Global methods} \citep{jiralerspong2024efficient,roy2025causal, 10966043,abdulaal2023causal,peng2026causcientist, vashishtha2025causal,takayama2025integrating}, which use LLMs to assess causal relations over all $O(n^2)$ variable pairs, and (2) the \textbf{Undirected-only method} \citep{long2023causal} which applies LLMs only to undirected edges in a pre-computed equivalence class.
A central challenge in current LLM-augmented causal discovery is deciding when LLM knowledge should override statistical evidence when neither source is known to be reliable. Statistical algorithms may reflect finite-sample errors, different inductive biases, or ambiguity in observational evidence, while LLM judgments can introduce incorrect causal relations. Without access to the ground-truth graph, the relative reliability of these sources cannot be directly measured. 

CauTion addresses this challenge through annotation-free reliability estimation and selective arbitration. It uses cross-algorithm consensus as proxy supervision, estimates the reliability of statistical and semantic evidence separately for edge existence and direction, and queries the LLM when its response can affect the calibrated voting decision. This design aims to exploit useful domain knowledge while limiting the damage caused by unreliable semantic judgments.
\sys operates in three stages.
(1) \emph{Algorithm ensemble} aggregates outputs from multiple causal discovery algorithms and resolves variable pairs on which all algorithms agree, thereby reducing individual algorithm bias and limiting LLM intervention to disputed cases.
(2) \emph{Trust-calibrated arbitration} estimates the reliability of the LLM and the algorithm ensemble, and resolves disputed edges through trust-weighted aggregation, querying the LLM only when algorithmic evidence is insufficiently reliable.
(3) \emph{Cycle repair} ensures that the final graph satisfies the DAG constraint.
Together, these stages allow \sys to use reliable algorithmic consensus, selectively invoke LLM domain knowledge for uncertain edges, and combine both sources through calibrated trust.
Figure~\ref{fig:intro} shows the advantages of \sys over existing LLM-augmented methods.

\sys outperforms all baselines in SHD and F1 during experiments, with gains over other LLM-augmented baselines becoming more significant as graph scale grows. On the largest benchmark Win95pts ($n{=}76$), \sys achieves an SHD (Structural Hamming Distance) of 28 compared to 69 for the second-best LLM-augmented baseline.
Moreover, \sys demonstrates strong robustness to LLM selection, maintaining superior performance across six LLMs of varying capability.

To summarize, our main contributions are:
\begin{itemize}[leftmargin=*]
    \item \emph{Conceptually}, we formulate LLM-augmented causal discovery as a trust-calibrated arbitration problem. This perspective treats LLMs and data-centric algorithms as complementary but imperfect evidence sources, and emphasizes that LLM knowledge should be incorporated only after its reliability is calibrated against an ensemble of statistical causal discovery algorithms.
    \item \emph{Methodologically}, we propose \sys, a trust-calibrated ensemble framework for causal discovery. It combines cross-algorithm consensus, reliability-aware LLM arbitration, and DAG-preserving cycle repair to construct reliable and valid causal graphs.
    \item \emph{Empirically}, we show that \sys consistently improves causal graph recovery across benchmarks and graph scales. \sys achieves stronger recovery performance while reducing unnecessary LLM queries, with gains becoming more pronounced as the number of variables increases.
\end{itemize}

\section{Related Work}
\label{sec:related_work}
\paragraph{Data-driven Methods}
Purely data-driven causal discovery methods rely on the statistical properties of observational data and span a range of paradigms, including constraint-based methods (e.g., PC~\citep{spirtes2000causation}, FCI~\citep{10.5555/2074158.2074215}), score-based methods (e.g., GES~\citep{chickering2002optimal}, GRaSP~\citep{lam2022grasp}, CaMML~\citep{Wallace1996CausalDV}), and continuous optimization methods (e.g., NOTEARS~\citep{zheng2018dags}, DAGMA~\citep{bello2022dagma}). Constraint-based methods are sensitive to test errors under finite samples and can only recover a Markov equivalence class (MEC), leaving some edge directions unresolved. 
Score-based methods optimize a scoring function over an exponentially large graph space, relying on greedy or stochastic 
search heuristics that cannot guarantee global optimality. Moreover, they depend on distributional assumptions in score functions that may not hold in practice.
Continuous optimization methods introduce non-convex objectives that are sensitive to hyperparameter choices and thresholding. 
Finally, these paradigms share a common limitation: without external domain knowledge, purely statistical methods cannot reliably resolve causal structure beyond what the data alone can identify.

\paragraph{LLM-Augmented Methods}
Existing LLM-augmented causal discovery methods apply LLMs without verifying their domain-specific reliability and reference on at most one data-driven CD algorithm. Therefore, most of them are sensitive to LLM errors and single algorithm bias. Based on how they involve LLMs in the discovery process, they can be broadly categorized into two categories: \textbf{global LLM methods} and \textbf{undirected only method}, as summarized in Figure \ref{fig:intro} and Appendix \ref{app:baselines} Table~\ref{tab:comparison}.
(1) The \textbf{global LLM methods} query the LLM across all variable pairs.
%
LLM-BFS \citep{jiralerspong2024efficient} and Causal-LLM 
\citep{roy2025causal} use LLM alone to generate full graphs.
ET-MCMC \citep{10966043} provides more reliability by restricting queries to the edges that the LLM reports high confidence in and post-processing the results with a CD algorithm. 
CMA \citep{abdulaal2023causal} establishes an iterative feedback loop between LLM proposals and a deep structural causal model~\citep{pawlowski2020deep}.
CauScientist \citep{peng2026causcientist} validates each LLM-proposed 
modification with a BIC score.
Most of these methods require the LLM to reason over $O(n^2)$ variable pairs in a single call, leading to degraded decision quality as graph size grows. While Causal Order \citep{vashishtha2025causal} and SCP \citep{takayama2025integrating} 
avoid this by querying one relationship at a time, they incur 
$O(kn^2)$ and $O(n^2)$ total queries respectively, leading to high token costs. The high decision volume makes global LLM methods difficult to balance decision quality and computational cost. 
%
(2) The \textbf{undirected only method} LLM-MEC \citep{long2023causal} reduces decision volume by 
restricting LLM queries to undirected edges within a pre-computed 
MEC. However, its sequential edge resolution causes errors to 
accumulate: each orientation is immediately propagated via Meek 
rules, so a single LLM error can trigger a cascade of forced 
orientations across the remaining undirected edges. 
A detailed analysis of each method is shown in Appendix \ref{app:baselines}.

\section{Methods}
\label{sec:method}
\begin{figure}[t]
  \centering
  \includegraphics[width=1\textwidth]{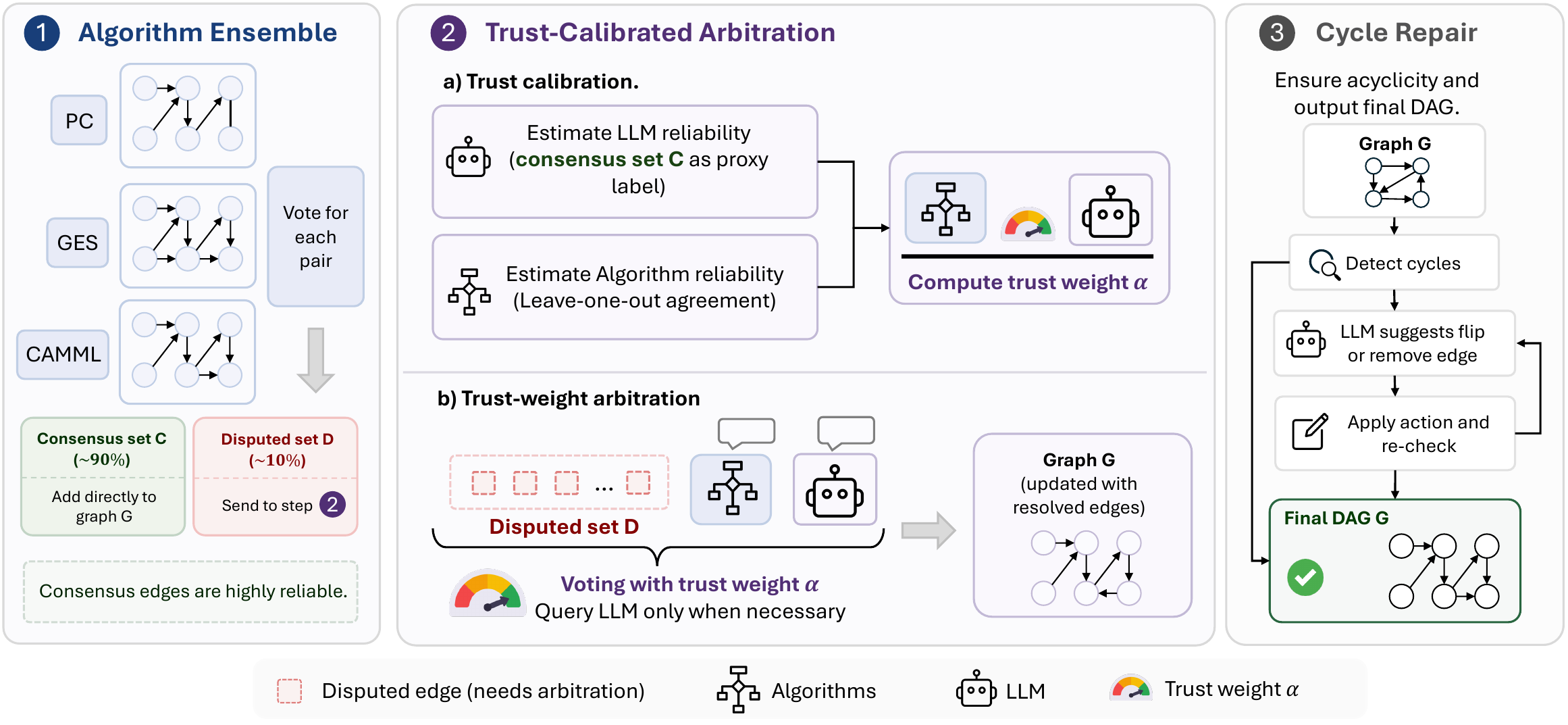}
  \caption{\textbf{Overview of \sys.} 1) We first aggregate outputs from multiple causal discovery algorithms (PC, GES, CAMML) and partition variable pairs into consensus and disputed sets. Consensus edges are resolved directly, leaving disputed edges for further arbitration. 2) We use the consensus set as proxy labels to estimate the reliability of LLM, while using leave-one-out to estimate the algorithms. Then compute the calibrated trust weight, which is used for LLM arbitration. 3) A final cycle-repair step ensures acyclicity of the output graph.}
  \label{fig:method}
\end{figure}

We propose \sys, a framework that reliably integrates calibrated LLM domain knowledge into an ensemble of statistical causal discovery algorithms. 
The full pipeline is illustrated in Figure~\ref{fig:method} and summarized in Algorithm~\ref{alg:main}.
We first formulate the problem setting in Section~\ref{problem_setup}.
We then introduce the three core components of \sys: algorithm ensemble for extracting cross-algorithm consensus (Section \ref{sec:consensus}), trust-calibrated arbitration for resolving disputed edges (Section \ref{arbitration}), and cycle repair for enforcing the DAG constraint (Section \ref{sec:cycle_repair}).

\subsection{Problem Setup}
\label{problem_setup}
Let $\mathcal{V} = \{X_1, \ldots, X_n\}$ be a set of $n$ variables, and let $\mathcal{A}$ denote a set of causal discovery algorithms. Each algorithm $A \in \mathcal{A}$ produces a partially directed graph over $\mathcal{V}$. For every unordered variable pair $e =\{i,j\}$, the output of algorithm $A$ is mapped to one of four vote categories:
\begin{equation*}
    v_A(e) \in \{\text{fwd}\ (i \to j),\ \text{rev}\ (j \to i),\ 
    \text{undir}\ (i - j),\ \text{none}\}
\end{equation*}
\noindent where \texttt{undir} indicates that the algorithm detected an undirected edge but could not orient it, and \texttt{none} indicates no edge. The goal is to aggregate these per-algorithm votes into a single directed acyclic graph (DAG).

\subsection{Algorithm Ensemble}
\label{sec:consensus}

For each variable pair $e$, we accumulate weighted votes across 
algorithms. Undirected votes contribute equally to both directions:
\begin{align*}
    w_\text{fwd}(e)  &= \textstyle\sum_{A \in \mathcal{A}} \bigl(
        \mathbf{1}[v_A = \text{fwd}] + 
        0.5 \cdot \mathbf{1}[v_A = \text{undir}]\bigr) \\
    w_\text{rev}(e)  = \textstyle\sum_{A \in \mathcal{A}} \bigl(
        \mathbf{1}[v_A &= \text{rev}]  + 
        0.5 \cdot \mathbf{1}[v_A = \text{undir}]\bigr) \quad
    w_\text{none}(e) = \textstyle\sum_{A \in \mathcal{A}} \mathbf{1}[v_A = \text{none}]
\end{align*}
A pair $e=\{i,j\}$ is resolved by consensus if any of the following conditions holds: $w_\text{fwd}(e) = |\mathcal{A}|$, $w_\text{rev}(e) = |\mathcal{A}|$, or $w_\text{none}(e) = |\mathcal{A}|$. All remaining pairs form the disputed set $\mathcal{D}$. Only the disputed set is passed to the resolution pipeline in Section~\ref{sec:calib}.

Consensus-based filtering serves a dual purpose: it reduces the number 
of pairs requiring LLM arbitration, and it yields high-confidence 
pseudo-ground-truth labels for downstream calibration. On the Win95pts dataset with $n=76$, 
consensus agreement is achieved on the majority of variable pairs (96.6\%), with near-perfect accuracy (99.6\%)—confirming their suitability as proxy labels.

\subsection{Trust-Calibrated Arbitration}
\label{arbitration}

\subsubsection{Estimating Source Reliability}
\label{sec:calib}

Before resolving disputed edges, we independently estimate the 
predictive accuracy of each algorithm and of the LLM directly from the 
data, requiring no external annotation. Consensus edges serve as 
proxy ground-truth labels for LLM calibration, and leave-one-out 
agreement serves as proxy labels for algorithm calibration. The 
resulting estimates are used to compute per-task trust weights that 
govern how much each source is relied upon during voting.

We first estimate the target distribution $(p_\text{exist}, p_\text{none})$ 
over $\mathcal{D}$, i.e.\ the fraction of disputed pairs that involve a 
directed edge versus no edge. Using the vote weights defined in 
Section~\ref{sec:consensus}, the soft edge fraction for each disputed 
pair and the aggregate target distribution are:
\begin{equation*}
    p_\text{exist}^{(e)} = 
        (w_\text{fwd}^{(e)} + w_\text{rev}^{(e)})/{|\mathcal{A}|},
    \qquad
    p_\text{exist} = \frac{1}{|\mathcal{D}|} 
    \textstyle \sum_{e \in \mathcal{D}} p_\text{exist}^{(e)},
    \qquad
    p_\text{none} = 1 - p_\text{exist}
\end{equation*}
This distribution is used to stratify LLM calibration samples and to 
reweight per-algorithm accuracy estimates to reflect the disputed-edge 
distribution, as described below.

All LLM queries share a three-stage pipeline: the LLM first infers a 
domain description from variable names, which is injected into all 
subsequent prompts; it then infers a semantic interpretation for each 
variable (e.g., \texttt{Dyspnoea [shortness of breath]}), which is 
appended inline to variable names in all edge-level queries. For 
each edge-level query, the LLM is presented with variable names and 
their interpretations. To mitigate position bias, 
direction queries are issued twice per pair---once in the $A \to B$ 
framing and once in $B \to A$---and only consistent answers are 
retained.

\subsubsection{LLM Calibration}
\label{sec:llm_calib}

We estimate LLM accuracy on consensus edges, assuming that the LLM's 
domain knowledge is uniformly distributed across consensus and disputed 
pairs---an assumption that holds when both sets are drawn from the same 
domain and variable space. We treat edge existence and direction as two 
independent tasks, since they reflect distinct reasoning capabilities: 
existence requires knowing whether a causal mechanism operates between 
two variables, while direction requires knowing which variable is the 
cause and which is the effect. Separate accuracy estimates are therefore 
maintained for each task, for both the LLM and the algorithm ensemble. We sample $n_\text{calib} = 30$ consensus 
edges stratified by the target distribution:
\begin{equation*}
    n_\text{calib,exist} = \mathrm{round}(n_\text{calib} \times p_\text{exist}),
    \qquad
    n_\text{calib,none} = n_\text{calib} - n_\text{calib,exist}
\end{equation*}
If the edge pool is smaller than $n_\text{calib,exist}$, the shortfall 
rolls over to the none pool.

\paragraph{Existence task.}
The LLM is queried for whether an edge exists and returns 
$\hat{e} \in \{\texttt{exist},\ \texttt{none}\}$. 
We compute per-class accuracy conditioned on the ground-truth label 
$d^*_e$: $acc_{\text{cond,exist}}$ is the fraction of ground-truth 
edge pairs correctly identified as \texttt{exist}, and 
$acc_{\text{cond,none}}$ is the fraction of ground-truth none pairs 
correctly identified as \texttt{none}. The effective existence accuracy 
is reweighted to the target distribution to remove sampling bias:
\begin{equation}
\label{eq:exist_reweight}
    acc_{\text{LLM,exist}} = 
    p_\text{exist} \cdot acc_{\text{cond,exist}} + 
    p_\text{none} \cdot acc_{\text{cond,none}}
\end{equation}

\paragraph{Direction task.}
Each edge-class pair is queried twice to eliminate position bias, as 
described above; only pairs where both queries return the same direction 
are counted as consistent. The direction accuracy is then:
$acc_{\text{LLM,dir}} = c_{\text{dir}} / n_{\text{dir}}$,
where $n_\text{dir}$ is the number of consistent pairs and $c_\text{dir}$ 
is the subset where the consistent answer matches $d^*_e$.

\subsubsection{Algorithm LOO Calibration}
\label{sec:algo_calib}

We estimate each algorithm's accuracy via leave-one-out (LOO) 
evaluation: for each pair $e$, algorithm $A$'s prediction is 
compared against the majority vote of all remaining algorithms, which 
serves as a proxy label, with tied edges excluded. For existence, we 
compute the fraction of proxy-exist pairs and proxy-none pairs correctly 
predicted by $A$, then reweight by $(p_\text{exist}, p_\text{none})$ 
following Equation~\ref{eq:exist_reweight}, yielding $acc_{\text{exist}}^A$. 
For direction, $acc_{\text{dir}}^A$ is the fraction of pairs correctly 
oriented by $A$ among those where $A$ votes a definite direction, 
excluding undirected votes. To aggregate across algorithms, we use the 
mean restricted to algorithms at or above the median:
\begin{equation*}
    acc_{\text{algo,exist}} = \text{AboveMedianMean}_A
    \bigl(acc_{\text{exist}}^A\bigr), \qquad 
    acc_{\text{algo,dir}} = \text{AboveMedianMean}_A
    \bigl(acc_{\text{dir}}^A\bigr)
\end{equation*}
Since poorly performing algorithms receive lower vote weights during 
aggregation, the ensemble accuracy is better reflected by the 
higher-performing half, rather than a mean pulled down by weak algorithms.

\subsubsection{Trust Weights}
\label{sec:trust}

Given the calibrated accuracies $acc_{\text{LLM,exist}}$, 
$acc_{\text{LLM,dir}}$, $acc_{\text{algo,exist}}$, and 
$acc_{\text{algo,dir}}$, we compute two independent trust weights 
$\alpha_\text{exist}$ and $\alpha_\text{dir}$ that quantify the relative 
reliability of the LLM versus the algorithm ensemble on each task. Since 
both tasks reduce to binary decisions, random performance corresponds to 
an accuracy of 0.5. Trust weights are computed as a power-ratio of 
above-chance accuracies with exponent $\beta > 1$ (default $\beta = 3$), 
which sharpens the contrast between sources of similar accuracy:
\begin{equation*}
\begin{aligned}
        \alpha_\text{exist} = 
    \frac{\max(0,\ acc_{\text{LLM,exist}} - 0.5)^\beta}
         {\max(0,\ acc_{\text{LLM,exist}} - 0.5)^\beta + 
          \max(0,\ acc_{\text{algo,exist}} - 0.5)^\beta},
    \\
    \alpha_\text{dir} = 
    \frac{\max(0,\ acc_{\text{LLM,dir}} - 0.5)^\beta}
         {\max(0,\ acc_{\text{LLM,dir}} - 0.5)^\beta +
          \max(0,\ acc_{\text{algo,dir}} - 0.5)^\beta}
\end{aligned}
\end{equation*}
A higher $\alpha$ indicates greater relative trust in the LLM. When 
both sources perform at or below chance, we fall back to $\alpha = 0.5$, 
treating both sources as equally unreliable.

\subsubsection{Calibrated Voting}
\label{sec:voting}

For each disputed pair $e \in \mathcal{D}$, we first determine 
whether an edge exists, then determine its direction.

\paragraph{Existence vote.}
Algorithm votes are aggregated into a weighted existence score:
\begin{equation*}
    v_{\text{exist}}(e) = \frac{\sum_{A \in \mathcal{A}} 
    acc_{\text{exist}}^A \cdot \mathbf{1}[v_A \in 
    \{\text{fwd, rev, undir}\}]}{\sum_{A \in \mathcal{A}} 
    acc_{\text{exist}}^A}, \qquad v_{\text{none}}(e) = 1 - v_{\text{exist}}
\end{equation*}
When the LLM is consulted and returns $\hat{e} \in 
\{\texttt{exist},\ \texttt{none}\}$, the combined score is:
\begin{equation*}
    \label{eq:edge_existence_final}
    q_{\text{exist}} = (1-\alpha_{\text{exist}}) \cdot v_{\text{exist}} 
    + \alpha_{\text{exist}} \cdot \mathbf{1}[\hat{e}=\texttt{exist}],
    \qquad
    q_{\text{none}} = 1 - q_{\text{exist}}
\end{equation*}
with decision $\arg\max(q_{\text{exist}},\ q_{\text{none}})$. In 
practice, the LLM is only queried when $m_{\text{exist}} = 
|v_{\text{exist}} - v_{\text{none}}| \leq \alpha_{\text{exist}} / 
(1 - \alpha_{\text{exist}})$, since beyond this margin the algorithm 
vote is decisive regardless of the LLM response. This further reduces 
the number of LLM queries.

\paragraph{Direction vote.}
If an edge is determined to exist, direction is resolved using votes 
from algorithms that predict a definite direction 
($v_A \in \{\text{fwd, rev}\}$):
\begin{equation*}
    v_{\text{fwd}}(e) = \frac{\sum_{A:\ v_A \in \{\text{fwd, rev}\}} 
    acc_{\text{dir}}^A \cdot \mathbf{1}[v_A = \text{fwd}]}
    {\sum_{A:\ v_A \in \{\text{fwd, rev}\}} acc_{\text{dir}}^A}, 
    \qquad v_{\text{rev}}(e) = 1 - v_{\text{fwd}}
\end{equation*}
The combined score incorporating the LLM response 
$\hat{d} \in \{\texttt{fwd},\ \texttt{rev}\}$ is:
\begin{equation*}
\label{eq:edge_direction_final}
    q_{\text{fwd}} = (1 - \alpha_{\text{dir}}) \cdot v_{\text{fwd}} 
    + \alpha_{\text{dir}} \cdot \mathbf{1}[\hat{d}=\texttt{fwd}],
    \qquad
    q_{\text{rev}} = 1 - q_{\text{fwd}}
\end{equation*}
with decision $\arg\max(q_{\text{fwd}},\ q_{\text{rev}})$. Similar to the 
existence vote, the LLM 
is queried only when the algorithm margin $m_{\text{dir}} = 
|v_{\text{fwd}} - v_{\text{rev}}|$ falls below the analogous threshold 
$\alpha_{\text{dir}} / (1 - \alpha_{\text{dir}})$. When queried, direction queries follow the 
double-framing procedure described above; if the two framings are inconsistent, the decision falls back to the algorithm vote alone.

\subsection{Cycle Repair}
\label{sec:cycle_repair}

The voting procedure above produces a directed graph but does not guarantee a valid DAG. We therefore apply a cycle repair step to further process the graph.
We first detect cycles in the produced graph. If cycles are detected, for each edge $e$ on a detected cycle, we compute the weighted existence and direction margins:
\begin{equation*}
    m_{\text{exist}}(e) = |v_{\text{exist}} - v_{\text{none}}|, \qquad
    m_{\text{dir}}(e)   = |v_{\text{fwd}}  - v_{\text{rev}}|
\end{equation*}
An edge $e$ is designated as a \textit{repair candidate} if it is algorithmically 
uncertain in either sense:
\begin{equation*}
    \text{candidate}(e) = \mathbf{1}\bigl[m_{\text{exist}}(e) < 
    \theta_{\text{exist}}\bigr] \;\vee\; 
    \mathbf{1}\bigl[m_{\text{dir}}(e) < \theta_{\text{dir}}\bigr]
\end{equation*}
where $\theta_{\text{exist}} = \alpha_{\text{exist}} / 
(1-\alpha_{\text{exist}})$ and $\theta_{\text{dir}} = 
\alpha_{\text{dir}} / (1-\alpha_{\text{dir}})$ are the same thresholds used in Section \ref{sec:voting}.

\textit{Repair candidates} are then presented to the LLM together with the cycle path 
and variable interpretations. For each candidate, the LLM may choose from one 
of two actions: \textsc{Flip} (reverse the edge direction, available 
when $m_{\text{dir}}(e) < \theta_{\text{dir}}$) or \textsc{Remove} (delete the edge, available 
when $m_{\text{exist}}(e) < \theta_{\text{exist}}$). Each selected 
action is applied immediately, and the action is removed from the \textit{available action} set. New cycle check is performed, and if cycles 
remain, the procedure repeats until all cycles have been removed or there is no \textit{available action} left.
If all \textit{available actions} are exhausted and cycles persist, edges on the 
remaining cycle are removed in ascending order of $v_{\text{exist}}$ 
until the graph is acyclic.
\section{Experiments}
\label{sec:experiments}

\subsection{Experimental Setup}
\label{sec:experimental_setup}

\paragraph{Datasets}
We evaluate on six datasets from the bnlearn repository~\citep{scutari2010learning}\footnote{\url{https://www.bnlearn.com/bnrepository/}}: Cancer ($n{=}5$, $|E|{=}4$)~\citep{korb2010bayesian}, Insurance ($n{=}27$, $|E|{=}52$)~\citep{binder1997adaptive}, Water ($n{=}32$, $|E|{=}66$)~\citep{jensen1989expert}, Alarm ($n{=}37$, $|E|{=}46$)~\citep{beinlich1989alarm}, Barley ($n{=}48$, $|E|{=}84$)~\citep{kristensen2002use}, and Win95pts ($n{=}76$, $|E|{=}112$)~\citep{heckerman1995decision}, spanning a range of scales and edge densities. For each dataset, we sample five versions of observations, with 5000 samples in each version.

\paragraph{Metrics} Three metrics were used. \textbf{Structural Hamming Distance (SHD)} counts the minimum number of edge insertions, deletions, and reversals required to transform the predicted graph into the ground truth, with lower values indicating better performance. \textbf{F1 score} measures the harmonic mean of precision and recall over directed edges. \textbf{Structural Intervention Distance (SID)}~\citep{peters2015structural} quantifies structural errors in intervention inference. It yields a single exact value for DAG outputs and lower and upper bounds for CPDAG outputs.

\paragraph{Baselines}
We compare against three categories of data-centric methods, including \textbf{constraint-based} PC~\citep{spirtes2000causation} and FCI~\citep{10.5555/2074158.2074215}, \textbf{score-based} GES~\citep{chickering2002optimal}, GRaSP~\citep{lam2022grasp}, and CAMML~\citep{Wallace1996CausalDV}, and \textbf{continuous optimization-based} NOTEARS-MLP~\citep{zheng2020learning} and DAGMA~\citep{bello2022dagma}. We compare four LLM-augmented methods, including \textbf{global} methods LLM-BFS~\citep{jiralerspong2024efficient}, 
SCP~\citep{takayama2025integrating}, ET-MCMC~\citep{10966043}, and \textbf{undirected-only} method
LLM-MEC~\citep{long2023causal}. We additionally include CORR-LLM, a simple baseline that prompts the LLM with pairwise Pearson correlations to generate the full causal graph in a single query.

\paragraph{LLM backend and majority vote algorithms}
We use \claude as the main LLM backend for all LLM-agumented methods including ours. We use PC, GES and CAMML as the algorithm ensemble for all experiments, selected for their efficiency (Table~\ref{tab:runtime}), wide adoption in the literature, and diversity in underlying search strategies.

\subsection{Performance}
\begin{table}[t]
\centering
\caption{Performance across six datasets (SHD $\downarrow$, SID $\downarrow$; F1 in \% $\uparrow$). Best values are in \textbf{bold}. “---” indicates that the exact SID value is unavailable. Results are averaged across 5 runs \citep{wu2025sample,peng2026causcale}. \sys achieves the best or tied-best SHD and F1 on all datasets.}
\label{tab:main_results}
{\small
\setlength{\tabcolsep}{0.8pt}
\begin{tabular}{l ccc| ccc| ccc| ccc| ccc| ccc}
\toprule
\textbf{Method}
& \multicolumn{3}{c}{Cancer} & \multicolumn{3}{c}{Insurance} & \multicolumn{3}{c}{Water} & \multicolumn{3}{c}{Alarm} & \multicolumn{3}{c}{Barley} & \multicolumn{3}{c}{Win95pts} \\
& \multicolumn{3}{c}{\small$(n{=}5)$}
& \multicolumn{3}{c}{\small$(n{=}27)$}
& \multicolumn{3}{c}{\small$(n{=}32)$}
& \multicolumn{3}{c}{\small$(n{=}37)$}
& \multicolumn{3}{c}{\small$(n{=}48)$}
& \multicolumn{3}{c}{\small$(n{=}76)$} \\
\cmidrule(lr){2-4}\cmidrule(lr){5-7}\cmidrule(lr){8-10}\cmidrule(lr){11-13}\cmidrule(lr){14-16}\cmidrule(lr){17-19}
& SHD & F1 & SID & SHD & F1 & SID & SHD & F1 & SID & SHD & F1 & SID & SHD & F1 & SID & SHD & F1 & SID \\
\midrule
\multicolumn{19}{l}{\textit{Constraint-based}}\\
PC & 2 & 50.0 & 10 & 25 & 63.7 & --- & 55 & 46.1 & --- & 14 & 85.9 & --- & 56 & 49.4 & --- & 72 & 56.6 & --- \\
FCI & 4 & 66.7 & 10 & 31 & 70.0 & --- & 59 & 46.8 & --- & 33 & 71.8 & --- & 67 & 60.2 & --- & 101 & 58.5 & --- \\
\midrule
\multicolumn{19}{l}{\textit{Score-based}}\\
GES & \textbf{0} & \textbf{100.0} & \textbf{0} & 22 & 72.9 & --- & 47 & 50.2 & --- & 8 & 91.2 & --- & 54 & 64.0 & --- & 40 & 80.8 & --- \\
GRaSP & 2 & 80.0 & --- & 41 & 63.3 & --- & 52 & 48.3 & --- & 13 & 87.1 & --- & 86 & 45.4 & --- & 58 & 73.4 & --- \\
CAMML & 2 & 50.3 & 9 & 27 & 60.5 & 340 & 55 & 31.8 & 567 & 9 & 85.8 & 96 & 77 & 34.7 & 1144 & 66 & 67.1 & 358 \\
\midrule
\multicolumn{19}{l}{\textit{Continuous optimization}}\\
NOTEARS-MLP & 3 & 26.7 & 10 & 50 & 33.8 & 535 & 63 & 9.4 & 539 & 52 & 30.6 & 427 & 84 & 19.3 & 1191 & 142 & 23.3 & 1010 \\
DAGMA & 4 & 0.0 & 10 & 51 & 27.0 & 587 & 58 & 20.9 & 522 & 33 & 58.0 & 252 & 82 & 5.6 & 1233 & 103 & 43.9 & 674 \\
\midrule
\multicolumn{19}{l}{\textit{LLM-augmented}}\\
CORR-LLM & \textbf{0} & \textbf{100.0} & \textbf{0} & 23 & 74.7 & 375 & 42 & 53.5 & 404 & 16 & 80.2 & 110 & 78 & 46.1 & 881 & 78 & 58.0 & 588 \\
LLM-BFS & \textbf{0} & \textbf{100.0} & \textbf{0} & 25 & 73.3 & 326 & 54 & 51.8 & \textbf{374} & 44 & 49.4 & 268 & 121 & 28.1 & 1023 & 147 & 23.2 & 1017 \\
SCP & 1 & 70.0 & 6 & 20 & 72.3 & --- & 53 & 47.3 & --- & 10 & 87.2 & --- & 57 & 47.0 & --- & 69 & 58.7 & --- \\
ET-MCMC & 0 & 91.4 & 1 & 21 & 73.7 & 285 & 52 & 44.5 & 525 & 10 & 86.5 & 63 & 90 & 29.0 & 1218 & 95 & 60.7 & 385 \\
LLM-MEC & 2 & 50.0 & 10 & 24 & 64.1 & --- & 54 & 44.7 & --- & 13 & 84.9 & --- & 61 & 44.3 & --- & 72 & 56.6 & --- \\
\midrule
\rowcolor{blue!8}
Ours & \textbf{0} & \textbf{100.0} & \textbf{0} & \textbf{13} & \textbf{85.7} & \textbf{199} & \textbf{39} & \textbf{59.2} & 377 & \textbf{5} & \textbf{93.8} & \textbf{38} & \textbf{43} & \textbf{68.4} & \textbf{743} & \textbf{28} & \textbf{84.2} & \textbf{343} \\
\bottomrule
\end{tabular}
}
\end{table}

Table~\ref{tab:main_results} reports results across six benchmarks. \textbf{\sys achieves the best or tied-best mean SHD and F1 on all six benchmarks.} Among the reported exact SID values, \sys also achieves the best or tied-best result on five benchmarks. Water is the exception: LLM-BFS achieves the lowest SID at 374, slightly below CauTion's 377, while CauTion achieves the best SHD and F1.

\textbf{\sys outperforms all statistical baselines on all metrics.} It mitigate algorithm bias by aggregating signals from multiple algorithms, and applies reliable LLM knowledge to compensate for what statistical algorithms cannot resolve alone.
\textbf{Among LLM-augmented baselines, existing methods scale poorly to larger graphs.} In \textbf{global LLM methods}, CORR-LLM, LLM-BFS, ET-MCMC performs well on small networks, but degrades sharply on larger graphs. Though SCP shows higher performance compared to the other three global methods, its token cost is 100 times larger than ours (Table \ref{tab:runtime}).
Though the \textbf{undirected only method} LLM-MEC shows more stable behavior and low token cost than global methods, its performance is constrained by its reliance on a single upstream algorithm (PC). Moreover, it incurs 8 $\times$ higher run time than \sys due to large computing complexity for MECs.
\sys addresses these failure modes by restricting LLM queries to disputed edges and downweighting LLM influence when its estimated domain accuracy is low, limiting exposure to unreliable LLM responses. The gains over LLM-augmented baselines become substantially more pronounced as graph scale grows, consistent with the framework design: larger graphs yield more edges, amplifying the value of consensus filtering and selective LLM integration.

To evaluate performance beyond published benchmarks, we additionally test \sys on Airport, a newly constructed synthetic Bayesian network with 15 variables and 23 directed edges. We compare \sys against the same baselines as in the main experiments. \sys achieves the best or tied-best performance across all three metrics on this network, with construction details and full results provided in Appendix~\ref{app:airport}.

\subsection{Ablation Study}
We evaluate the contribution of each component in our pipeline by 
comparing the following ablated strategies: \emph{Consensus only}, 
which accepts consensus edges and discards all disputed edges; 
\emph{LLM Only}, which predicts all variable pairs using the LLM alone; \emph{Majority Vote}, which resolves each disputed edge by selecting the option with the most votes across algorithms; \emph{Cons + LLM}, which resolves all disputed edges using LLM responses alone.

\begin{figure}[t]
  \centering
  \includegraphics[width=1\textwidth]{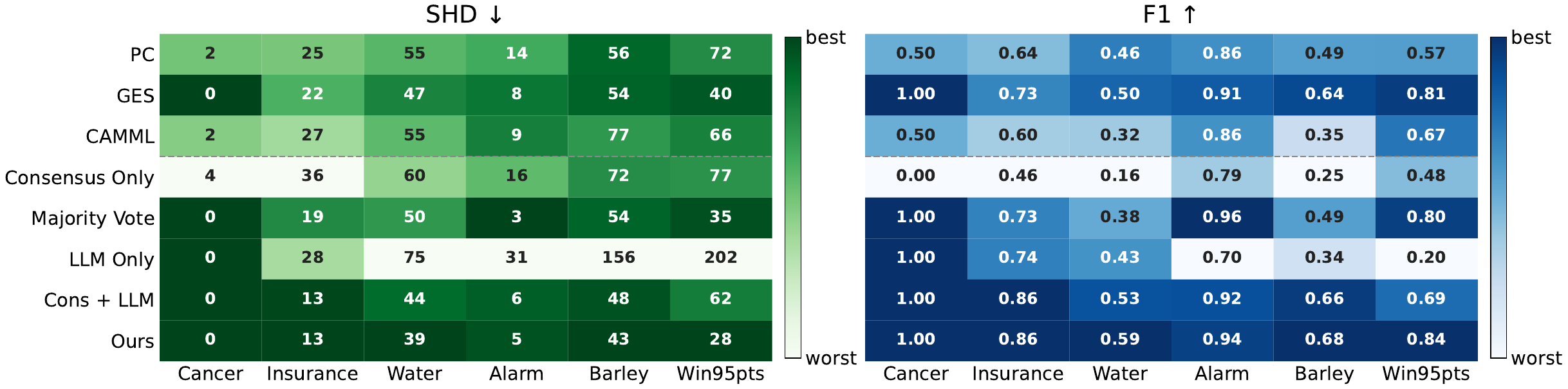}
  \caption{Ablation study with various edge decision strategies. Color is normalized column-wise. The three algorithms used in ensemble aggregation are included above the dashed line for reference.}
  \label{fig:heatmap}
\end{figure}

Shown in Figure~\ref{fig:heatmap}, several trends emerge. First, selecting \emph{Consensus only} is not enough for causal discovery, leaving many edges unresolved. \emph{Majority Vote} outperforms individual algorithms, confirming that \textbf{ensemble aggregation mitigates single algorithm bias}. However, without LLM domain knowledge, it remains constrained by the inherent limitations of purely statistical inference. On the other hand, \emph{LLM Only} achieves competitive performance on small graphs utilizing domain knowledge but degrades sharply at larger scale graph. \emph{Cons + LLM} demonstrates that \textbf{combining algorithmic consensus with LLM knowledge substantially reduces LLM error exposure}. However, its suboptimal performance on Alarm, and Win95pts compared to 
\emph{Majority Vote} suggests that LLM reliability on disputed edges is overestimated. \sys outperforms both \emph{Majority Vote} and \emph{Cons + LLM} on disputed edges, demonstrating that \textbf{calibrated trust weighting effectively balances the respective strengths of the algorithm ensemble and LLM domain knowledge}.

\subsection{Robustness to LLM Model Selection}

\begin{figure}[t]
  \centering
  \includegraphics[width=1\textwidth]{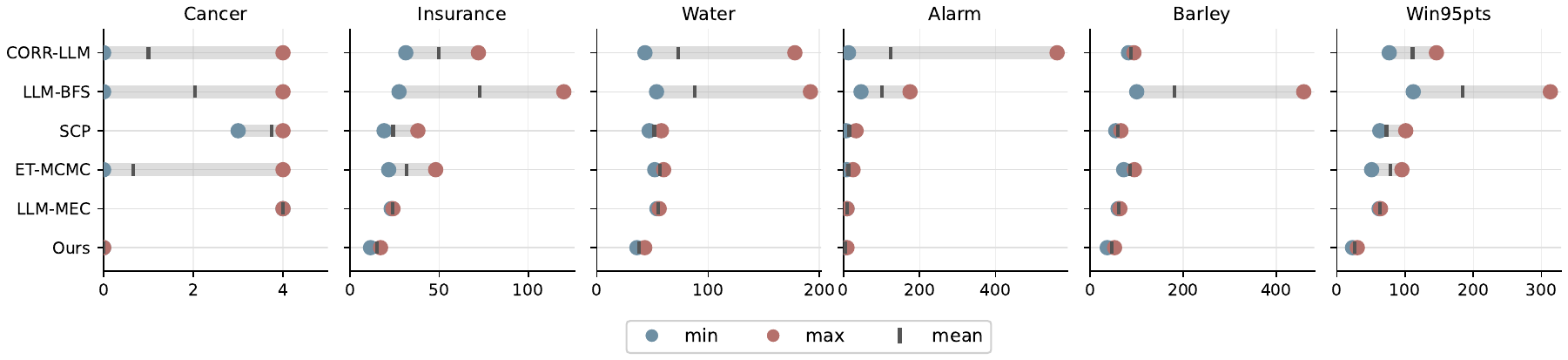}
  \caption{Cross-model robustness of LLM-augmented methods in terms of SHD. Each horizontal bar spans the minimum to maximum SHD achieved across six LLM backends, where shorter bars indicate greater robustness to backend selection.
  }
  \label{fig:robustness-shd}
\end{figure}

We further evaluate robustness to LLM backend selection. We test six LLMs with varying levels of capability: \claude, GPT-5.2, Llama-3.3-70B-Instruct, DeepSeek-V3.2, Qwen3-8B, and Llama-3.1-8B-Instruct, covering both strong and weak models to test sensitivity to LLM errors. Figure~\ref{fig:robustness-shd} reports the SHD range across models for each LLM-augmented method and dataset.
\textbf{\sys achieves the lowest SHD on all six datasets with low performance fluctuation across LLM backends, demonstrating robustness to LLM errors.}
Among \textbf{global LLM methods}, fully LLM-driven methods (CORR-LLM, LLM-BFS) exhibit severe sensitivity on large graphs. CORR-LLM reaches SHD 562 on Alarm for the worst backend (Llama-3.1-8B-Instruct) compared to 13 for the best. LLM-BFS similarly ranges from 100 to 459 on Barley, with Qwen3-8B achieving 100 and DeepSeek-V3.2 collapses to 459, indicating that LLM methods alone are fragile in practice causal discovery. SCP and ET-MCMC are more stable than CORR-LLM and LLM-BFS, but still exhibit substantially higher cross-model variance and consistently worse performance than \sys. \textbf{The undirected-only method} LLM-MEC shows little cross-model variance, but remains bottlenecked by its single upstream algorithm (PC). More detailed comparisons in terms of SHD and F1 are provided in Appendix~\ref{app:robust} (Figures~\ref{fig:app-robust-shd}--\ref{fig:app-robust-f1}), where x-axes are zoomed in per dataset to better highlight cross-model variance. Figures~\ref{fig:app-radar-shd} and~\ref{fig:app-radar-f1} further provide radar plots showing the performance of all backbend models across datasets and methods.

\section{Conclusion}
\label{sec:conclusion_impact}
\label{sec:limitation}
In conclusion, we present \sys, a trust-calibrated ensemble framework for LLM-augmented causal discovery. By combining cross-algorithm consensus filtering, annotation-free reliability estimation, and calibrated trust-weighted voting, \sys addresses two fundamental limitations of existing approaches: the unverified reliability of LLM domain knowledge and the vulnerability of single-algorithm methods to algorithm-specific bias.
Experiments across six benchmarks of varying scale demonstrate that \sys consistently outperforms all baselines in SHD and F1.
\clearpage

\subsection*{AI use statement}
In this work, we used generative AI tools for polish writing.
We have not used generative AI tools for retrieval and discovery, research ideation or execution, draft sections of the paper, generating synthetic datasets, and providing mathematical claims are not applicable to this work.
We have reviewed all AI-assisted work. We reviewed the AI-edited text for clarity and coherence and revised it where necessary. We take responsibility for the final content of this work,
including text, claims or artifacts produced with the aid of generative AI.

\subsection*{Reproducibility statement}

To facilitate reproducibility, we describe the \sys method in Section~\ref{sec:method} and provide the full algorithmic pseudocode in Appendix~\ref{app:method_details}. The LLM prompt templates and query settings are provided in Appendix~\ref{sec:prompting}. Section~\ref{sec:experimental_setup} specifies the benchmarks, sample sizes, LLM backend, and ensemble algorithms, while Appendices~\ref{app:dataset_details}--\ref{sec:baseline_implementation} detail data generation, evaluation metrics, and baseline implementations. Appendix~\ref{app:exp-performance} reports results across independently sampled dataset versions, together with the hardware setup, runtime, and token usage. An anonymized implementation is available at \url{https://anonymous.4open.science/r/CauTion-37C6}.

\bibliography{iclr2027_conference}
\bibliographystyle{iclr2027_conference}

\clearpage
\appendix
\section{Related Work Details}
\label{app:baselines}
\subsection{Data-driven Baselines}
\subsubsection{Constraint-Based Methods}

\paragraph{PC \citep{spirtes2000causation}.}
PC is a landmark constraint-based algorithm that recovers causal 
structure via two phases: skeleton discovery through conditional 
independence (CI) tests, followed by orientation of v-structures and 
propagation via Meek rules. 
PC outputs a Completed Partially Directed Acyclic Graph (CPDAG) 
representing the MEC of the true DAG, leaving edges whose directions 
cannot be determined from observational data unoriented. It is sensitive to CI test errors under finite samples during skeleton discovery phase,
and its errors propagate through the orientation phase. 

\paragraph{FCI \citep{10.5555/2074158.2074215}.}
FCI extends PC to handle latent confounders and selection bias, outputting Partial Ancestral Graph (PAG) rather than CPDAG. FCI introduces additional edge marks (circle, tail, arrowhead) that encode uncertainty about ancestral relationships. It inherits PC's sensitivity to CI test errors and error propagation.

\subsection{Score-Based Methods}

Unless otherwise noted, methods in this category search over equivalence classes and output a CPDAG, leaving some edge directions unresolved from observational data alone.

\paragraph{GES \citep{chickering2002optimal}.}
GES performs a two-phase greedy search over the space of CPDAGs: a forward phase that adds edges to maximize a decomposable score (e.g., BIC, BDeu), followed by a backward phase that removes edges. Its performance is sensitive to the choice of score function (usually BIC or BDeu) and susceptible to local optima due to its greedy nature.

\paragraph{GRaSP \citep{lam2022grasp}.}
GRaSP is a permutation-based method that leverages the Sparsest Permutation (SP) principle, searching for the causal ordering that yields the sparsest DAG under a given score. It improves upon earlier SP-based methods in scalability but remains computationally expensive on dense graphs. 

\paragraph{CAMML \citep{Wallace1996CausalDV}.}
CAMML is a Bayesian score-based method that uses the Minimum Message Length (MML) principle to search over DAGs via MCMC. Unlike the above methods, CAMML outputs a fully oriented DAG rather than a CPDAG. However, its MCMC search scales poorly to large graphs: the DAG space grows super-exponentially with the number of variables, making it difficult to achieve adequate mixing and convergence in practice. 

\subsection{Continuous Optimization Methods}

\paragraph{NOTEARS-MLP \citep{zheng2020learning}.}
NOTEARS-MLP extends the original NOTEARS framework \citep{zheng2018dags} to non-linear relationships by parameterizing each variable's conditional distribution with a multilayer perceptron (MLP). The acyclicity constraint is enforced via a smooth algebraic characterization, converting the combinatorial graph search into a continuous optimization problem. While this allows gradient-based optimization, the resulting objective is highly non-convex due to both the non-linear acyclicity constraint and the MLP parameterization, making the optimization prone to local optima and sensitive to initialization. In practice, the method requires careful tuning of the regularization strength and the thresholding applied to continuous edge weights to recover a discrete graph structure. 

\paragraph{DAGMA \citep{bello2022dagma}.}
DAGMA replaces the NOTEARS acyclicity constraint with a 
log-determinant-based characterization that avoids numerical 
instability in the original formulation. While more stable than 
NOTEARS in optimization, DAGMA shares the same fundamental 
limitations: non-convex objectives, hyperparameter sensitivity, 
and reliance on thresholding to obtain a discrete graph from 
continuous edge weights. 

\begin{table*}[t]
\centering
\small
\setlength{\tabcolsep}{5pt}
\begin{tabular}{lccccc}
\toprule
\textbf{Method}
  & \makecell{\textbf{Decision} \\ \textbf{Volume}}
  & \makecell{\textbf{LLM Error} \\ \textbf{Exposure}}
  & \makecell{\textbf{Algorithmic} \\ \textbf{Bias Exposure}}
  & \makecell{\textbf{Uncertainty} \\ \textbf{Quantification}}
  & \textbf{DAG} \\
\hline
\multicolumn{6}{c}{\cellcolor{green!10}\emph{Category I: Global LLM Methods}}\\
\hline
LLM-BFS
  & $O(n^2)$ & High & None  & \ding{55} & \ding{51} \\
Causal-LLM
  & $O(n^2)$ & High & None  & \ding{55} & \ding{51} \\
Causal Order
  & $O(kn^2)$ & High & High & \ding{55} & \ding{51} \\
SCP
  & $O(n^2)$ & High & High & \ding{55} & \ding{55} \\
ET-MCMC
  & $O(n^2)$ & High & High & \ding{55} & \ding{51} \\
CMA
  & $O(Tn^2)$ & High & High & \ding{55} & \ding{55} \\
CauScientist
  & $O(Tn^2)$ & Medium & High & \ding{55} & \ding{51} \\
\hline
\multicolumn{6}{c}{\cellcolor{teal!10}\emph{Category II: Undirected Only Methods}}\\
\hline
LLM-MEC
  & Undirected only & Medium & High & \ding{55} & \ding{55} \\
\hline
\multicolumn{6}{c}{\cellcolor{orange!10}\emph{Category III: Ensemble Calibrated (Ours)}}\\
\hline
\textbf{\sys~(Ours)}
  & Disputed only & Low & Low & \ding{51} & \ding{51} \\
\bottomrule
\end{tabular}%
\caption{Comparison of LLM-integrated causal discovery methods.
\textit{Decision Volume}: the total number of edges that the LLM is required to analyze during causal structure learning.
\textit{LLM Error Exposure}: how much the final graph is affected by incorrect LLM predictions.
\textit{Algorithmic Bias Exposure}: degree to which the final graph is subject to errors from a single statistical algorithm. None indicates no statistical algorithm is used.
\textit{Uncertainty Quantification}: whether LLM reliability is explicitly estimated from data.
\textit{DAG}: whether the output graph is guaranteed to be a DAG.}
\label{tab:comparison}
\end{table*}

\subsection{LLM-Augmented Baselines}

\paragraph{LLM-BFS \citep{jiralerspong2024efficient}.}
LLM-BFS constructs the causal graph via a Breadth-First Search 
(BFS) over the variable space, querying the LLM over all 
variable pairs with a total decision volume of $O(n^2)$. 
The procedure begins by asking the LLM to identify root 
variables---those not caused by any other variable in the 
system. Starting from these roots, the BFS iteratively expands each 
node by prompting the LLM to select all variables directly 
caused by it, visiting each node exactly once, resulting in 
$O(n)$ calls of $O(n)$ per-call complexity. Since all edges are determined solely by LLM responses 
without any statistical grounding, errors in root node identification or downstream variable selection propagate 
through the entire traversal without any correction mechanism.

\paragraph{Causal-LLM \citep{roy2025causal}.}
Causal-LLM offers two operating modes depending on data 
availability. When variable metadata is accessible, it uses 
a prompt-based approach to generate the full causal graph in 
a single LLM call via in-context learning, with a total 
decision volume of $O(n^2)$, achieved through $O(1)$ calls 
of $O(n^2)$ per-call complexity. When metadata is absent, 
it falls back to a purely data-driven approach using LLM 
representations. The two modes operate independently. Asking the LLM to jointly 
reason over all variable pairs in a single prompt places a 
heavy burden on the model's context window, leading to 
degraded decision quality on larger graphs with no 
statistical safeguard against erroneous outputs.

\paragraph{Causal Order \citep{vashishtha2025causal}.}
Causal Order infers causal ordering priors by querying the 
LLM with variable triplets, with a total decision volume of 
$O(kn^2)$ achieved through $O(kn^2)$ calls of $O(1)$ 
per-call complexity, where $k$ is the number of triplets 
per variable pair. The inferred ordering is then used to 
guide a downstream statistical algorithm. The high query 
complexity limits its scalability to small graphs only (maximum 
$n{=}24$ in the original paper). Since the ordering prior 
influences all downstream edge decisions, both LLM error 
exposure and algorithmic bias exposure cover the full graph: 
errors in the LLM-inferred ordering propagate into the 
downstream algorithm, whose own biases further affect all 
edges without any calibration to verify whether the LLM's 
ordering judgments are reliable.

\paragraph{SCP \citep{takayama2025integrating}.}
SCP first runs a statistical algorithm (e.g., PC) to obtain an initial 
graph, then provides the result as context to the LLM via 
``Statistical Causal Prompting,'' querying it on each of 
the $O(n^2)$ variable pairs to produce a probability matrix, 
which is further processed by a downstream statistical 
algorithm. The total decision volume is $O(n^2)$, achieved 
through $O(n^2)$ calls of $O(1)$ per-call complexity. Since 
the LLM is queried over all variable pairs, LLM error 
exposure covers the entire graph. Furthermore, both the 
upstream and downstream statistical algorithms are single 
algorithms, making the final graph susceptible to the 
individual algorithm limitations and failure modes.

\paragraph{ET-MCMC \citep{10966043}.}
ET-MCMC first queries the LLM on all $O(n^2)$ variable pairs 
to assess confidence, then uses only the high-confidence 
subset as soft ancestral constraints for a downstream 
score-based method. The total decision volume is $O(n^2)$, 
achieved through $O(n^2)$ calls of $O(1)$ per-call 
complexity. Although LLM queries are restricted to high-confidence pairs with greater reliability, 
empirical studies have shown that LLMs tend to be overconfident \citep{xiong2024can}. LLM 
errors can still affect any edge in the final graph unchecked.
The final graph is further subject to the individual limitations and biases of 
the single downstream algorithm.

\paragraph{CMA \citep{abdulaal2023causal}.}
CMA combines LLM metadata-based reasoning with Deep Structural 
Causal Models (DSCMs) in an iterative agentic loop. The LLM 
proposes graph modifications in both global and local phases, 
with the DSCM providing data-driven feedback at each step. 
The total decision volume is $O(Tn^2)$ across $T$ iterations, 
as the LLM operates over the full graph at each round. Since 
all edges are subject to LLM-proposed modifications at every iteration, LLM error exposure covers the entire graph. Moreover, the method is subject to the limitations and assumptions of the single DSCM used for validation in each iteration.

\paragraph{CauScientist \citep{peng2026causcientist}.}
CauScientist operates in three stages: hybrid initialization selects the better of a data-driven baseline graph and an 
LLM-proposed graph based on BIC score; iterative refinement where the LLM proposes atomic edge modifications given the current full graph as context, evaluated by a BIC-based 
verifier; and an error memory that prevents the LLM from 
repeating previously rejected proposals. The total decision 
volume is $O(Tn^2)$ across $T$ iterations, achieved through 
$T$ calls each of $O(n^2)$ per-call complexity as the full 
graph is provided as context at every step. LLM error 
exposure is partially mitigated by the BIC verifier, but BIC-based validation provides only a local structural criterion that may not be reliable under finite samples.

\paragraph{LLM-MEC \citep{long2023causal}.}
LLM-MEC restricts LLM queries to the undirected edges within a pre-computed MEC, iteratively orienting edges with minimum risk of eliminating the true graph from the MEC. The total decision volume is $|\mathcal{U}|$, the number of undirected 
edges in the MEC, achieved through $|\mathcal{U}|$ calls of $O(1)$ per-call complexity. Since only undirected edges are queried, LLM error exposure is limited to these edges. However, the method relies on a single upstream algorithm (e.g., PC) to compute the initial MEC, making the entire graph vulnerable to the individual limitations of that algorithm. 
Beyond this, the method resolves edges sequentially: each orientation decision is immediately propagated via Meek rules 
to infer additional edge directions before the next edge is 
queried. This means a single LLM error can trigger a cascade 
of forced orientations, amplifying the impact of individual 
mistakes across the remaining undirected edges.
Moreover, the LLM error rate used in the risk computation is derived 
from human assumption or LLM token probabilities rather 
than empirical calibration, which can deviate substantially 
from the true error rate. Finally, the iteration 
stops when the cumulative risk of excluding the true graph 
exceeds a threshold, so the output is not guaranteed to be 
a fully oriented DAG---some edges may remain undirected in 
the final graph.

\section{Method Details}
\label{app:method_details}
Algorithm~\ref{alg:main} summarizes the complete \sys pipeline, which proceeds in three stages: consensus-based pre-filtering, accuracy calibration, and calibrated voting with cycle repair.
\begin{pseudocode}{\sys: Calibrated LLM-Augmented Causal Discovery}{alg:main}
\begin{algorithmic}[1]

\Require Dataset $\mathcal{X}$, variable names $\mathcal{V}$, 
         algorithm set $\mathcal{A}$, LLM backend, 
         calibration budget $n_\text{calib}$, 
         trust exponent $\beta$
\Ensure  DAG $G$ over $\mathcal{V}$

\vspace{4pt}
\State \textbf{// Stage 1: Consensus Pre-Filtering}
\State Initialize $G \gets (\mathcal{V},\emptyset)$
\For{each $A \in \mathcal{A}$}
    \State Run $A$ on $\mathcal{X}$ $\to$ vote matrix $\{v_A(e)\}$ 
           for all pairs $e$
\EndFor
\State Partition all pairs into $\mathcal{C}$ (consensus) and 
       $\mathcal{D}$ (disputed) \hfill $\triangleright$ \S\ref{sec:consensus}
\State Add unanimously supported directed edges from $\mathcal{C}$ to $G$

\vspace{4pt}
\State \textbf{// Stage 2: Accuracy Calibration}
\State Estimate target distribution 
       $(p_\text{exist},\, p_\text{none})$ from $\mathcal{D}$
\State Infer domain context and variable semantics via LLM 
       \hfill $\triangleright$ \S\ref{sec:prompting}

\State Sample up to $n_\text{calib}$ pairs from $\mathcal{C}$
       proportional to $(p_\text{exist}, p_\text{none})$
\State Query LLM on samples $\to$ 
       $acc_\text{LLM,exist}$,\; $acc_\text{LLM,dir}$ 
       \hfill $\triangleright$ \S\ref{sec:calib}

\State Compute LOO accuracies $acc_{\text{exist}}^A$ and
       $acc_{\text{dir}}^A$ for each $A \in \mathcal{A}$
       on $\mathcal{C} \cup \mathcal{D}$
       \hfill $\triangleright$ \S\ref{sec:algo_calib}
\State Aggregate $\to$
       $acc_{\text{algo,exist}}$, \;
       $acc_{\text{algo,dir}}$

\State Compute trust weights $\alpha_\text{exist}$, $\alpha_\text{dir}$ 
       from accuracies \S\ref{sec:trust}

\vspace{4pt}
\State \textbf{// Stage 3: Calibrated Voting on Disputed Pairs}
\State Compute algorithm vote scores and margins for all pairs
       \hfill $\triangleright$ \S\ref{sec:voting}
\For{each $e \in \mathcal{D}$}
    \State $q_\text{exist} \gets v_\text{exist}(e)$
    \If{$m_\text{exist}(e) \leq \theta_\text{exist}$}
        \State Query LLM $\to \hat{e} \in \{\texttt{exist},\texttt{none}\}$
        \State $q_\text{exist} \gets (1-\alpha_\text{exist}) \cdot 
               v_\text{exist}(e) + \alpha_\text{exist} \cdot
               \mathbf{1}[\hat{e}=\texttt{exist}]$
    \EndIf
    \State $q_\text{none} \gets 1-q_\text{exist}$
    \If{$q_\text{none} \geq q_\text{exist}$}
        \State \textbf{continue} \Comment{No edge added}
    \EndIf
    \State $q_\text{fwd} \gets v_\text{fwd}(e)$
    \If{$m_\text{dir}(e) < \theta_\text{dir}$}
        \State Query LLM with both variable orders; map answers to the same order
        \If{both answers agree on $d_\text{LLM} \in \{\texttt{fwd},\texttt{rev}\}$}
            \State $q_\text{fwd} \gets \alpha_\text{dir} \cdot 
                   \mathbf{1}[d_\text{LLM}=\texttt{fwd}] +
                   (1-\alpha_\text{dir}) \cdot v_\text{fwd}(e)$
        \EndIf
    \EndIf
    \State $q_\text{rev} \gets 1-q_\text{fwd}$;
           $\hat{d} \gets \arg\max_{d\in\{\texttt{fwd},\texttt{rev}\}} q_d$
    \State Add directed edge to $G$ according to $\hat{d}$
\EndFor

\vspace{4pt}
\State \textbf{// Stage 4: Cycle Repair}
\State $\mathcal{T} \gets \emptyset$
\While{$G$ contains a directed cycle $\mathcal{C}_\text{cyc}$}
    \State $\mathcal{S} \gets$ eligible actions on $\mathcal{C}_\text{cyc}$
           excluding $\mathcal{T}$ \hfill $\triangleright$ \S\ref{sec:cycle_repair}
    \If{$\mathcal{S}\neq\emptyset$}
        \State Query LLM to select one action $a \in \mathcal{S}$
        \State Apply $a$ to $G$; $\mathcal{T} \gets \mathcal{T}\cup\{a\}$
    \Else
        \State Remove edge with lowest $v_\text{exist}$ 
               from $\mathcal{C}_\text{cyc}$ \Comment{Fallback}
    \EndIf
\EndWhile

\State \Return $G$

\end{algorithmic}
\end{pseudocode}

\clearpage
\subsection{LLM Prompting}
\label{sec:prompting}

All LLM queries follow a shared prompting structure detailed below.
Placeholders in \texttt{\{braces\}} denote runtime-substituted values;
their contents are described in Table~\ref{tab:placeholders}.
\textit{Italic} annotations in the prompts are explanatory comments and are
not part of the actual prompt text.

\begin{table}[h]
\centering
\small
\caption{Placeholder variables used across all prompts.}
\label{tab:placeholders}
\begin{tabular}{ll}
\toprule
\textbf{Placeholder} & \textbf{Description} \\
\midrule
\texttt{\{domain\_name\}}     & Domain label (e.g.\ ``Alarm'') \\
\texttt{\{domain\_context\}}  & Domain description inferred by the LLM \\
\texttt{\{variables\}}        & Variable list with categorical values and meanings\\
\texttt{\{N\}}        & Number of variables \\
\texttt{\{consensus\_directed\_edges\}} & Directed edges agreed upon by all algorithms \\
\texttt{\{pairs\}}            & Candidate variable pairs to be queried \\
\texttt{\{cycle\}}            & Detected directed cycle path (e.g.\ $A \to B \to C \to A$) \\
\texttt{\{candidates\}}       & Uncertain edges on the cycle with their algorithmic margins \\
\bottomrule
\end{tabular}
\end{table}

We use four categories of prompts. \textbf{Domain inference (P1)} is called
once at the start to infer a domain label and description from variable names,
which is injected into the system prompt (P2) of all subsequent calls.
\textbf{Variable interpretation (P3)} is called once before before querying disputed edges to
assign a natural-language meaning to each variable;
during calibration, consensus edges are withheld to prevent label leakage.
\textbf{Edge-level queries (P4-5)} cover the the existence and direction queries issued during calibration and disputed-edge resolution; direction queries are issued twice per pair in opposite framings
to eliminate position bias, and only consistent answers are retained.
\textbf{Cycle repair} (P6) is called when a directed cycle is detected in the
assembled graph, presenting uncertain cycle edges to the LLM for correction
via \textsc{Flip} or \textsc{Remove} actions.

LLM calls use a temperature of 0.0 to ensure deterministic outputs. Edge existence and direction queries are batched into groups of at most 10 variable pairs per call to keep prompts within a manageable length.

\subsubsection*{P1. Domain Inference Prompt}

\begin{tcolorbox}[colback=gray!8, colframe=gray!40, left=4pt, right=4pt, top=2pt, bottom=2pt]
{\small\sffamily\bfseries System}\\[4pt]
{\small\ttfamily You are analyzing a system to support causal reasoning about its variables.}
\end{tcolorbox}

\begin{tcolorbox}[colback=gray!8, colframe=gray!40, left=4pt, right=4pt, top=2pt, bottom=2pt]
{\small\sffamily\bfseries User}\\[4pt]
{\small\ttfamily
Here are variables from an unknown system:\\[4pt]
\{variables\} {\normalfont\small\itshape (one line per variable: "~~- NAME: val1, val2, ...")}\\[4pt]
What is this system about?}
\end{tcolorbox}

\subsubsection*{P2. Shared System Prompt}

\begin{tcolorbox}[colback=gray!8, colframe=gray!40, left=4pt, right=4pt, top=2pt, bottom=2pt]
{\small\sffamily\bfseries System}\\[4pt]
{\small\ttfamily
You are an expert in causal inference and the \{domain\_name\} domain.\\
Domain context: \{domain\_context\}\\
Your task is to assess the most plausible causal relationship between pairs of variables using your domain knowledge alone.}
\end{tcolorbox}

\subsubsection*{P3. Variable Interpretation Prompt}

\begin{tcolorbox}[colback=gray!8, colframe=gray!40, left=4pt, right=4pt, top=2pt, bottom=2pt]
{\small\sffamily\bfseries User}\\[4pt]
{\small\ttfamily
Before analyzing causal directions, build an initial understanding of what each variable in this dataset likely represents.

\medskip
Variables (\{N\} total):\\
\{variables\} {\normalfont\small\itshape (with optional value list)}

\medskip
=== Relationships agreed on by ALL algorithms ===\\
\{consensus\_directed\_edges\} {\normalfont\small\itshape (e.g. "~~VAR1 -\textgreater{} VAR2")}

\medskip
For each variable, describe what it likely measures or represents.
Rate your confidence from 0 to 10 (10 = completely certain, 0 = no idea).

\medskip
Output valid JSON only:\\
\{\\
\quad "variable\_interpretations": \{\\
\qquad "VAR\_NAME": \{\\
\qquad\quad "meaning": "...",\\
\qquad\quad "confidence": 8\\
\qquad \},\\
\qquad ...\\
\quad \}\\
\}}
\end{tcolorbox}

\subsubsection*{P4. Edge Existence Query Prompt}

\begin{tcolorbox}[colback=gray!8, colframe=gray!40, left=4pt, right=4pt, top=2pt, bottom=2pt]
{\small\sffamily\bfseries User}\\[4pt]
{\small\ttfamily
For each variable pair below, determine whether a direct causal relationship exists between them, using your domain knowledge only.

\medskip
Definitions:\\
\quad "EDGE" = a direct causal link exists between the two variables (in either direction)\\
\quad "NONE" = no direct causal link exists between them

\medskip
Variable meanings:\\
\{variables\} {\normalfont\small\itshape (e.g. "~~VAR1: meaning~~[values: val1, val2]")}

\medskip
Output a single JSON object. Each key is a pair number; each value has "reasoning" (brief explanation) and "existence" ("EDGE" or "NONE").

Example: \{"1": \{"reasoning": "...", "existence": "EDGE"\},

\phantom{Example: }"2": \{"reasoning": "...", "existence": "NONE"\}\}

\medskip
--- Variable pairs ---\\
\{pairs\} {\normalfont\small\itshape (e.g. "~~"1": VAR1 \textless{}-\textgreater{} VAR2")}}
\end{tcolorbox}

\subsubsection*{P5. Edge Direction Query Prompt}

\begin{tcolorbox}[colback=gray!8, colframe=gray!40, left=4pt, right=4pt, top=2pt, bottom=2pt]
{\small\sffamily\bfseries User}\\[4pt]
{\small\ttfamily
A direct causal edge is KNOWN to exist for each pair below.
Your task is ONLY to determine its direction.
You MUST output exactly "FWD" or "REV" for every pair --- no other values.

\medskip
Definitions:\\
\quad "FWD" = the first-listed variable directly causes the second\\
\quad "REV" = the second-listed variable directly causes the first

\medskip
Variable meanings:\\
\{variables\} {\normalfont\small\itshape (e.g. "~~VAR1: meaning~~[values: val1, val2]")}

\medskip
Output a single JSON object. Each key is a pair number; each value has "reasoning" (brief explanation) and "direction" ("FWD" or "REV").

Example: \{"1": \{"reasoning": "...", "direction": "FWD"\},

\phantom{Example: }"2": \{"reasoning": "...", "direction": "REV"\}\}

\medskip
--- Pairs to orient ---\\
\{pairs\} {\normalfont\small\itshape (e.g. "~~"1": VAR1 \textless{}-\textgreater{} VAR2~~(FWD = VAR1 -\textgreater{} VAR2~~|~~REV = VAR2 -\textgreater{} VAR1)")}}
\end{tcolorbox}

\subsubsection*{P6. Cycle Repair Prompt}

\begin{tcolorbox}[colback=gray!8, colframe=gray!40, left=4pt, right=4pt, top=2pt, bottom=2pt]
{\small\sffamily\bfseries User}\\[4pt]
{\small\ttfamily
A directed cycle was detected in the causal graph:

\quad \{cycle\} {\normalfont\small\itshape (e.g. "VAR1 -\textgreater{} VAR2 -\textgreater{} VAR3 -\textgreater{} VAR1")}

\medskip
Causal graphs must be acyclic (DAG). The cycle above needs to be corrected.

\medskip
Variable meanings:\\
\{variables\} {\normalfont\small\itshape (meanings for variables involved in the cycle)}

\medskip
The following candidate edges have uncertain existence or direction
(based on algorithm confidence). Review them and specify corrections
for those you think are incorrect:\\
\{candidates\}\\
{\normalfont\small\itshape (e.g. "~~"1": VAR1 -\textgreater{} VAR2~~(existence uncertain; margin=0.12)~~[actions: FLIP/REMOVE]")}

\medskip
Actions:\\
\quad "FLIP"~~~-- reverse the direction of the edge\\
\quad "REMOVE" -- delete this causal relationship entirely\\
\qquad\qquad\qquad\quad (only for candidates marked [actions: FLIP/REMOVE])

\medskip
Output ONLY a JSON object with a "reasoning" field and a "corrections" field.
"corrections" maps candidate number to action ("FLIP" or "REMOVE");
include only the edges you want to correct.

Example: \{"reasoning": "...", "corrections": \{"2": "FLIP"\}\}}
\end{tcolorbox}

\section{Experiment Details}
\label{app:experiment_details}
\subsection{Dataset Details}
\label{app:dataset_details}

The six main benchmarks are standard discrete Bayesian networks drawn from the \emph{bnlearn} repository~\citep{scutari2010learning}.
Each dataset is generated by forward-sampling from the ground-truth network specified by its \textsc{bif} file, using the \texttt{pgmpy} library~\citep{ankan2024pgmpy} (\texttt{BayesianModelSampling.forward\_sample}).
We additionally evaluate on a newly constructed synthetic network, \emph{Airport}, in Appendix~\ref{app:airport}.

\subsection{Metrics}
\label{sec:metrics}
We evaluate the performance of all causal discovery methods using the following structural metrics.

\paragraph{Structural Hamming Distance (SHD).} SHD counts the minimum number of edge insertions, deletions, and reversals needed to transform the predicted graph into the ground truth. 

\paragraph{Precision, Recall, and F1 Score.}
Delegated to \texttt{sklearn.metrics.f1\_score} and related functions~\citep{scikit-learn}, applied to the flattened adjacency matrices. Precision and Recall are computed at the edge level over the full $n \times n$ binary matrix; reversed edges thus count simultaneously as a false positive and a false negative.

\paragraph{Structural Interventional Distance (SID).}
We use the official R implementation of SID~\citep{peters2015structural}, which returns an exact value for DAG estimates and lower and upper bounds for CPDAG estimates. We report exact SID values as scalars and distinct lower and upper bounds as intervals $[L,U]$. Coincident bounds are also displayed as a single value. SID measures how useful the predicted graph actually is for causal inference. Concretely, the algorithm iterates over all ordered variable pairs $(i,j)$ and checks whether the predicted graph yields the same causal effect of $i$ on $j$ as the ground-truth graph. Each mismatch is counted as one inference error. SID is the total number of such mismatches across all variable pairs.

\subsection{Baseline Implementation Detail}
\label{sec:baseline_implementation}
\subsubsection{Data-driven Baselines}
PC~\citep{spirtes2000causation}, FCI~\citep{10.5555/2074158.2074215}, 
GES~\citep{chickering2002optimal}, and 
GRaSP~\citep{lam2022grasp} are implemented via 
\texttt{causal-learn}\footnote{\url{https://causal-learn.readthedocs.io/en/latest/}}~\citep{zheng2024causal}, 
using the chi-squared independence test and BDeu score for discrete data. 
CAMML~\citep{Wallace1996CausalDV} is implemented via the official 
BI-CaMML Java 
implementation\footnote{\url{https://bayesian-intelligence.com/software/BI-CaMML-Quickstart-Guide-1.4/}}.

NOTEARS-MLP~\citep{zheng2020learning} is implemented via the 
\texttt{gCastle} 
package\footnote{\url{https://gcastle-test.readthedocs.io/en/latest/getting_started.html}}~\citep{zhang2021gcastle}, 
with $z$-score standardization applied to the input data. 
DAGMA~\citep{bello2022dagma} is implemented via the official 
\texttt{dagma} Python 
package\footnote{\url{https://dagma.readthedocs.io/en/latest/}}.

\subsubsection{LLM-Augmented Baselines}

\paragraph{CORR-LLM.}
A zero-shot single-call baseline designed and implemented by us. The Pearson correlation matrix of the observations is 
passed to the prompt alongside variable names. The LLM is asked to output a an \texttt{edges} list. 
Post-processing: up to 3 retries if the returned graph contains cycles. If cycles persist after all candidate actions are exhausted, edges on the remaining cycle are randomly removed until the graph becomes acyclic.

\paragraph{LLM-BFS \citep{jiralerspong2024efficient}.}
We use the official \texttt{causal-llm-bfs} repository.

\paragraph{SCP \citep{takayama2025integrating}.}
We use the official repository for SCP.

\paragraph{ET-MCMC \citep{10966043}.}
We use the official repository for ET-MCMC.

\paragraph{LLM-MEC \citep{long2023causal}.}
We use the official repository for LLM-MEC.  Since not all LLMs support token-level log-probabilities, we handle two cases: for models that support logprobs, we follow the original implementation and extract $\log P(\text{A})$ and $\log P(\text{B})$ directly from the top token candidates; for models that do not  support logprobs, we fall back to parsing the text response and assigning a fixed probability of $(0.85, 0.15)$ to the chosen and unchosen options, respectively.
  
\subsection{Performance Detail}
\label{app:exp-performance}

\subsubsection{Full Results with Standard Deviations}
\label{app:sid_bounds}
\label{app:full_results}
We report mean\,$\pm$\,sample standard deviation across 5 independently sampled dataset versions. F1, precision, and recall are percentages.
For DAG outputs, SID is reported as mean\,$\pm$\,sample standard deviation. For CPDAG outputs, SID is characterized by lower and upper bounds corresponding to the minimum and maximum SID values among DAGs in the represented Markov equivalence class. For outputs with distinct SID bounds, we report $[\bar{L}\pm s_L,\bar{U}\pm s_U]$, with the mean and sample standard deviation computed separately for each bound across the five dataset versions.

\begin{table}[h]
\centering
{\small
\caption{Full results on \textbf{Cancer} ($n=5$, $|E|=4$).}
\label{tab:appendix_cancer}
\setlength{\tabcolsep}{4pt}
\begin{tabular}{l ccccc}
\toprule
\textbf{Method} & \textbf{SHD} & \textbf{F1 (\%)} & \textbf{Prec (\%)} & \textbf{Rec (\%)} & \textbf{SID} \\
\midrule
\multicolumn{6}{@{}l}{\textit{Constraint-based}}\\
PC & 2.0$\pm$0.0 & 50.0$\pm$0.0 & 50.0$\pm$0.0 & 50.0$\pm$0.0 & 10.0$\pm$0.0 \\
FCI & 4.0$\pm$0.0 & 66.7$\pm$0.0 & 50.0$\pm$0.0 & \textbf{100.0$\pm$0.0} & 10.0$\pm$0.0 \\
\midrule
\multicolumn{6}{@{}l}{\textit{Score-based}}\\
GES & \textbf{0.0$\pm$0.0} & \textbf{100.0$\pm$0.0} & \textbf{100.0$\pm$0.0} & \textbf{100.0$\pm$0.0} & 0.0$\pm$0.0 \\
GRaSP & 2.4$\pm$2.2 & 80.0$\pm$18.3 & 70.0$\pm$27.4 & \textbf{100.0$\pm$0.0} & [3.0$\pm$2.7, 9.0$\pm$8.2] \\
CAMML & 2.2$\pm$0.4 & 50.3$\pm$4.5 & 51.3$\pm$9.6 & 50.0$\pm$0.0 & 9.0$\pm$1.7 \\
\midrule
\multicolumn{6}{@{}l}{\textit{Continuous optimization}}\\
NOTEARS-MLP & 3.2$\pm$0.4 & 26.7$\pm$14.9 & 40.0$\pm$22.4 & 20.0$\pm$11.2 & 10.2$\pm$0.4 \\
DAGMA & 4.0$\pm$0.0 & 0.0$\pm$0.0 & 0.0$\pm$0.0 & 0.0$\pm$0.0 & 10.0$\pm$0.0 \\
\midrule
\multicolumn{6}{@{}l}{\textit{LLM-augmented}}\\
CORR-LLM & \textbf{0.0$\pm$0.0} & \textbf{100.0$\pm$0.0} & \textbf{100.0$\pm$0.0} & \textbf{100.0$\pm$0.0} & 0.0$\pm$0.0 \\
LLM-BFS & \textbf{0.0$\pm$0.0} & \textbf{100.0$\pm$0.0} & \textbf{100.0$\pm$0.0} & \textbf{100.0$\pm$0.0} & 0.0$\pm$0.0 \\
SCP & 1.2$\pm$0.4 & 70.0$\pm$11.2 & 70.0$\pm$11.2 & 70.0$\pm$11.2 & 6.0$\pm$2.2 \\
ET-MCMC & 0.4$\pm$0.9 & 91.4$\pm$19.2 & 93.3$\pm$14.9 & 90.0$\pm$22.4 & 1.2$\pm$2.7 \\
LLM-MEC & 2.0$\pm$0.0 & 50.0$\pm$0.0 & 50.0$\pm$0.0 & 50.0$\pm$0.0 & 10.0$\pm$0.0 \\
\midrule
\rowcolor{blue!8}
Ours & \textbf{0.0$\pm$0.0} & \textbf{100.0$\pm$0.0} & \textbf{100.0$\pm$0.0} & \textbf{100.0$\pm$0.0} & 0.0$\pm$0.0 \\
\bottomrule
\end{tabular}
}
\end{table}

\begin{table}[htbp]
\centering
{\small
\caption{Full results on \textbf{Insurance} ($n=27$, $|E|=52$).}
\label{tab:appendix_insurance}
\setlength{\tabcolsep}{4pt}
\begin{tabular}{l ccccc}
\toprule
\textbf{Method} & \textbf{SHD} & \textbf{F1 (\%)} & \textbf{Prec (\%)} & \textbf{Rec (\%)} & \textbf{SID} \\
\midrule
\multicolumn{6}{@{}l}{\textit{Constraint-based}}\\
PC & 24.8$\pm$1.5 & 63.7$\pm$3.1 & 71.7$\pm$4.1 & 57.3$\pm$2.5 & [398.6$\pm$15.1, 399.6$\pm$15.1] \\
FCI & 31.2$\pm$2.8 & 70.0$\pm$1.7 & 69.3$\pm$3.5 & 70.8$\pm$1.6 & [350.2$\pm$21.3, 426.6$\pm$20.8] \\
\midrule
\multicolumn{6}{@{}l}{\textit{Score-based}}\\
GES & 22.2$\pm$5.4 & 72.9$\pm$6.3 & 76.4$\pm$7.1 & 69.6$\pm$5.7 & [256.6$\pm$28.7, 333.2$\pm$30.9] \\
GRaSP & 40.6$\pm$10.0 & 63.3$\pm$6.4 & 59.4$\pm$10.1 & 68.5$\pm$3.5 & [285.6$\pm$33.9, 474.4$\pm$86.6] \\
CAMML & 27.4$\pm$6.9 & 60.5$\pm$10.6 & 65.6$\pm$11.8 & 56.2$\pm$9.7 & 340.4$\pm$26.5 \\
\midrule
\multicolumn{6}{@{}l}{\textit{Continuous optimization}}\\
NOTEARS-MLP & 50.2$\pm$5.2 & 33.8$\pm$6.6 & 40.8$\pm$8.1 & 28.8$\pm$5.6 & 535.2$\pm$54.3 \\
DAGMA & 51.0$\pm$3.8 & 27.0$\pm$4.6 & 36.5$\pm$8.2 & 21.5$\pm$3.2 & 587.4$\pm$31.5 \\
\midrule
\multicolumn{6}{@{}l}{\textit{LLM-augmented}}\\
CORR-LLM & 22.8$\pm$3.6 & 74.7$\pm$4.4 & 85.9$\pm$2.7 & 66.2$\pm$5.5 & 374.8$\pm$57.7 \\
LLM-BFS & 25.2$\pm$3.1 & 73.3$\pm$3.5 & 80.8$\pm$4.2 & 67.3$\pm$4.9 & 326.2$\pm$62.4 \\
SCP & 20.4$\pm$1.7 & 72.3$\pm$2.4 & 82.0$\pm$3.7 & 64.6$\pm$1.7 & [365.8$\pm$25.2, 397.8$\pm$20.1] \\
ET-MCMC & 21.0$\pm$5.3 & 73.6$\pm$6.3 & 76.4$\pm$8.1 & 71.2$\pm$4.9 & 285.0$\pm$24.7 \\
LLM-MEC & 24.2$\pm$1.1 & 64.1$\pm$2.8 & 72.7$\pm$3.4 & 57.3$\pm$2.5 & [398.6$\pm$15.1, 399.6$\pm$15.1] \\
\midrule
\rowcolor{blue!8}
Ours & \textbf{13.0$\pm$1.7} & \textbf{85.7$\pm$1.6} & \textbf{96.7$\pm$3.5} & \textbf{76.9$\pm$1.4} & 198.6$\pm$12.4 \\
\bottomrule
\end{tabular}
}
\end{table}

\begin{table}[htbp]
\centering
{\small
\caption{Full results on \textbf{Water} ($n=32$, $|E|=66$).}
\label{tab:appendix_water}
\setlength{\tabcolsep}{4pt}
\begin{tabular}{l ccccc}
\toprule
\textbf{Method} & \textbf{SHD} & \textbf{F1 (\%)} & \textbf{Prec (\%)} & \textbf{Rec (\%)} & \textbf{SID} \\
\midrule
\multicolumn{6}{@{}l}{\textit{Constraint-based}}\\
PC & 54.6$\pm$3.0 & 46.1$\pm$3.6 & 61.5$\pm$5.6 & 37.0$\pm$3.0 & [411.4$\pm$38.9, 531.8$\pm$43.1] \\
FCI & 59.2$\pm$3.3 & 46.8$\pm$2.2 & 57.2$\pm$4.7 & 39.7$\pm$1.3 & [386.8$\pm$12.7, 523.0$\pm$20.4] \\
\midrule
\multicolumn{6}{@{}l}{\textit{Score-based}}\\
GES & 47.0$\pm$2.5 & 50.2$\pm$2.8 & 69.1$\pm$3.6 & 39.4$\pm$2.4 & [423.0$\pm$15.1, 481.0$\pm$15.1] \\
GRaSP & 51.6$\pm$1.3 & 48.3$\pm$2.0 & 63.3$\pm$2.3 & 39.1$\pm$2.2 & [415.4$\pm$28.6, 523.2$\pm$31.8] \\
CAMML & 55.0$\pm$4.7 & 31.8$\pm$7.2 & 45.0$\pm$10.3 & 24.5$\pm$5.6 & 566.8$\pm$32.8 \\
\midrule
\multicolumn{6}{@{}l}{\textit{Continuous optimization}}\\
NOTEARS-MLP & 63.2$\pm$1.1 & 9.4$\pm$2.8 & 20.7$\pm$5.1 & 6.1$\pm$1.9 & 539.4$\pm$21.2 \\
DAGMA & 58.2$\pm$0.8 & 20.9$\pm$2.1 & 52.4$\pm$4.4 & 13.0$\pm$1.4 & 521.6$\pm$8.1 \\
\midrule
\multicolumn{6}{@{}l}{\textit{LLM-augmented}}\\
CORR-LLM & 42.0$\pm$0.0 & 53.5$\pm$0.5 & \textbf{99.2$\pm$1.7} & 36.7$\pm$0.7 & 403.8$\pm$2.7 \\
LLM-BFS & 54.0$\pm$17.0 & 51.8$\pm$14.2 & 64.2$\pm$19.5 & 43.6$\pm$11.9 & 374.4$\pm$24.0 \\
SCP & 52.6$\pm$4.1 & 47.4$\pm$3.2 & 65.2$\pm$7.0 & 37.3$\pm$2.3 & [405.4$\pm$26.1, 512.2$\pm$47.0] \\
ET-MCMC & 52.0$\pm$4.2 & 44.5$\pm$4.5 & 55.9$\pm$5.0 & 37.0$\pm$4.2 & 525.0$\pm$39.0 \\
LLM-MEC & 54.0$\pm$3.9 & 44.7$\pm$7.2 & 62.9$\pm$6.8 & 34.9$\pm$6.8 & [416.2$\pm$37.1, 519.6$\pm$38.6] \\
\midrule
\rowcolor{blue!8}
Ours & \textbf{39.0$\pm$4.3} & \textbf{59.2$\pm$6.2} & 90.6$\pm$9.1 & \textbf{43.9$\pm$4.8} & 377.0$\pm$55.7 \\
\bottomrule
\end{tabular}
}
\end{table}

\begin{table}[htbp]
\centering
{\small
\caption{Full results on \textbf{ALARM} ($n=37$, $|E|=46$).}
\label{tab:appendix_alarm}
\setlength{\tabcolsep}{4pt}
\begin{tabular}{l ccccc}
\toprule
\textbf{Method} & \textbf{SHD} & \textbf{F1 (\%)} & \textbf{Prec (\%)} & \textbf{Rec (\%)} & \textbf{SID} \\
\midrule
\multicolumn{6}{@{}l}{\textit{Constraint-based}}\\
PC & 13.6$\pm$5.4 & 85.9$\pm$5.1 & 82.4$\pm$8.0 & 90.0$\pm$1.9 & [44.8$\pm$12.6, 133.2$\pm$64.3] \\
FCI & 32.8$\pm$6.6 & 71.8$\pm$4.4 & 59.9$\pm$5.6 & 90.0$\pm$1.9 & [39.8$\pm$14.3, 170.4$\pm$82.4] \\
\midrule
\multicolumn{6}{@{}l}{\textit{Score-based}}\\
GES & 7.8$\pm$3.0 & 91.2$\pm$3.9 & 88.6$\pm$4.9 & 93.9$\pm$2.8 & [28.8$\pm$15.8, 77.6$\pm$21.4] \\
GRaSP & 12.8$\pm$6.3 & 87.1$\pm$5.7 & 81.2$\pm$9.2 & \textbf{94.3$\pm$1.9} & [31.8$\pm$21.8, 125.0$\pm$69.3] \\
CAMML & 8.8$\pm$3.4 & 85.8$\pm$6.0 & 85.5$\pm$7.0 & 86.1$\pm$5.0 & 95.8$\pm$51.6 \\
\midrule
\multicolumn{6}{@{}l}{\textit{Continuous optimization}}\\
NOTEARS-MLP & 52.2$\pm$4.2 & 30.6$\pm$3.8 & 27.3$\pm$4.1 & 34.8$\pm$3.4 & 427.0$\pm$56.6 \\
DAGMA & 33.4$\pm$6.9 & 58.0$\pm$7.9 & 56.6$\pm$8.4 & 59.6$\pm$7.8 & 251.6$\pm$26.4 \\
\midrule
\multicolumn{6}{@{}l}{\textit{LLM-augmented}}\\
CORR-LLM & 16.4$\pm$2.3 & 80.2$\pm$4.3 & 80.9$\pm$1.8 & 79.6$\pm$7.0 & 110.4$\pm$36.2 \\
LLM-BFS & 44.2$\pm$3.8 & 49.4$\pm$2.7 & 48.2$\pm$4.8 & 50.9$\pm$1.9 & 268.0$\pm$18.4 \\
SCP & 10.4$\pm$2.3 & 87.2$\pm$2.8 & 87.2$\pm$4.2 & 87.4$\pm$2.8 & [77.0$\pm$20.4, 115.0$\pm$28.6] \\
ET-MCMC & 10.2$\pm$4.6 & 86.5$\pm$6.1 & 83.8$\pm$8.3 & 89.6$\pm$3.6 & 63.4$\pm$45.0 \\
LLM-MEC & 13.0$\pm$4.8 & 84.9$\pm$6.8 & 82.7$\pm$8.3 & 87.4$\pm$5.4 & [71.0$\pm$48.4, 133.2$\pm$64.3] \\
\midrule
\rowcolor{blue!8}
Ours & \textbf{4.6$\pm$1.1} & \textbf{93.8$\pm$1.4} & \textbf{95.5$\pm$2.2} & 92.2$\pm$1.2 & 37.8$\pm$17.0 \\
\bottomrule
\end{tabular}
}
\end{table}

\begin{table}[htbp]
\centering
{\small
\caption{Full results on \textbf{Barley} ($n=48$, $|E|=84$).}
\label{tab:appendix_barley}
\setlength{\tabcolsep}{4pt}
\begin{tabular}{l ccccc}
\toprule
\textbf{Method} & \textbf{SHD} & \textbf{F1 (\%)} & \textbf{Prec (\%)} & \textbf{Rec (\%)} & \textbf{SID} \\
\midrule
\multicolumn{6}{@{}l}{\textit{Constraint-based}}\\
PC & 55.8$\pm$3.6 & 49.4$\pm$3.3 & 57.3$\pm$4.1 & 43.6$\pm$3.6 & [1012.2$\pm$115.6, 1087.6$\pm$68.2] \\
FCI & 67.4$\pm$6.5 & 60.2$\pm$3.0 & 58.1$\pm$4.1 & 62.6$\pm$2.0 & [740.8$\pm$55.0, 1110.0$\pm$47.3] \\
\midrule
\multicolumn{6}{@{}l}{\textit{Score-based}}\\
GES & 54.0$\pm$0.7 & 64.0$\pm$0.4 & 66.8$\pm$0.7 & 61.4$\pm$0.7 & [665.0$\pm$0.0, 1015.8$\pm$20.0] \\
GRaSP & 85.6$\pm$5.8 & 45.4$\pm$4.7 & 45.2$\pm$4.1 & 45.7$\pm$5.7 & [856.8$\pm$106.7, 1213.0$\pm$92.1] \\
CAMML & 77.2$\pm$10.2 & 34.7$\pm$9.7 & 38.7$\pm$10.6 & 31.4$\pm$8.9 & 1144.4$\pm$73.8 \\
\midrule
\multicolumn{6}{@{}l}{\textit{Continuous optimization}}\\
NOTEARS-MLP & 84.2$\pm$4.0 & 19.3$\pm$4.2 & 37.3$\pm$11.2 & 13.1$\pm$2.7 & 1190.6$\pm$67.9 \\
DAGMA & 82.2$\pm$1.3 & 5.6$\pm$2.9 & 30.1$\pm$13.7 & 3.1$\pm$1.6 & 1233.4$\pm$18.6 \\
\midrule
\multicolumn{6}{@{}l}{\textit{LLM-augmented}}\\
CORR-LLM & 78.2$\pm$5.4 & 46.1$\pm$4.1 & 54.0$\pm$4.2 & 40.2$\pm$4.2 & 881.2$\pm$62.5 \\
LLM-BFS & 120.8$\pm$10.4 & 28.1$\pm$5.1 & 28.2$\pm$1.5 & 29.0$\pm$8.9 & 1022.6$\pm$168.0 \\
SCP & 56.8$\pm$2.3 & 47.0$\pm$3.6 & 54.9$\pm$3.7 & 41.2$\pm$3.6 & [1069.8$\pm$114.6, 1132.6$\pm$72.6] \\
ET-MCMC & 90.2$\pm$7.0 & 29.0$\pm$2.7 & 31.0$\pm$4.0 & 27.4$\pm$2.1 & 1217.8$\pm$69.3 \\
LLM-MEC & 61.2$\pm$7.4 & 44.3$\pm$8.3 & 60.7$\pm$5.8 & 35.7$\pm$9.9 & [1019.4$\pm$104.8, 1094.2$\pm$67.7] \\
\midrule
\rowcolor{blue!8}
Ours & \textbf{42.8$\pm$5.0} & \textbf{68.4$\pm$3.7} & \textbf{74.8$\pm$7.0} & \textbf{63.3$\pm$4.3} & 743.4$\pm$101.6 \\
\bottomrule
\end{tabular}
}
\end{table}

\begin{table}[htbp]
\centering
{\small
\caption{Full results on \textbf{Win95pts} ($n=76$, $|E|=112$).}
\label{tab:appendix_win95pts}
\setlength{\tabcolsep}{4pt}
\begin{tabular}{l ccccc}
\toprule
\textbf{Method} & \textbf{SHD} & \textbf{F1 (\%)} & \textbf{Prec (\%)} & \textbf{Rec (\%)} & \textbf{SID} \\
\midrule
\multicolumn{6}{@{}l}{\textit{Constraint-based}}\\
PC & 72.0$\pm$5.2 & 56.7$\pm$3.0 & 64.8$\pm$4.3 & 50.4$\pm$2.2 & [819.8$\pm$52.9, 976.6$\pm$76.8] \\
FCI & 100.6$\pm$3.0 & 58.5$\pm$1.0 & 54.4$\pm$1.2 & 63.2$\pm$1.0 & [477.2$\pm$11.4, 967.2$\pm$65.4] \\
\midrule
\multicolumn{6}{@{}l}{\textit{Score-based}}\\
GES & 39.8$\pm$2.6 & 80.8$\pm$1.5 & 78.1$\pm$1.3 & \textbf{83.6$\pm$2.1} & [179.4$\pm$17.7, 230.2$\pm$12.7] \\
GRaSP & 57.8$\pm$16.1 & 73.4$\pm$6.6 & 70.4$\pm$8.0 & 76.8$\pm$5.1 & [320.0$\pm$134.9, 418.2$\pm$166.1] \\
CAMML & 65.6$\pm$11.4 & 67.1$\pm$5.3 & 63.9$\pm$6.5 & 70.7$\pm$4.8 & 358.2$\pm$46.2 \\
\midrule
\multicolumn{6}{@{}l}{\textit{Continuous optimization}}\\
NOTEARS-MLP & 141.6$\pm$12.1 & 23.3$\pm$4.6 & 21.6$\pm$4.3 & 25.4$\pm$5.0 & 1009.6$\pm$78.7 \\
DAGMA & 103.2$\pm$7.2 & 43.9$\pm$4.2 & 44.5$\pm$4.0 & 43.4$\pm$4.6 & 674.4$\pm$68.8 \\
\midrule
\multicolumn{6}{@{}l}{\textit{LLM-augmented}}\\
CORR-LLM & 77.6$\pm$7.5 & 58.0$\pm$3.7 & 70.4$\pm$4.3 & 49.3$\pm$3.6 & 587.8$\pm$51.0 \\
LLM-BFS & 146.6$\pm$8.5 & 23.2$\pm$3.3 & 27.3$\pm$4.2 & 20.2$\pm$2.8 & 1017.0$\pm$54.9 \\
SCP & 69.0$\pm$5.5 & 58.7$\pm$3.4 & 67.7$\pm$4.8 & 51.8$\pm$2.5 & [773.4$\pm$58.4, 922.0$\pm$85.8] \\
ET-MCMC & 94.6$\pm$10.5 & 60.7$\pm$4.5 & 52.0$\pm$4.4 & 72.9$\pm$4.7 & 384.8$\pm$76.0 \\
LLM-MEC & 71.8$\pm$5.5 & 56.6$\pm$2.9 & 64.9$\pm$4.4 & 50.2$\pm$2.0 & [823.8$\pm$46.6, 975.0$\pm$78.6] \\
\midrule
\rowcolor{blue!8}
Ours & \textbf{28.4$\pm$5.5} & \textbf{84.2$\pm$2.6} & \textbf{92.7$\pm$2.4} & 77.1$\pm$3.7 & 342.6$\pm$39.2 \\
\bottomrule
\end{tabular}
}
\end{table}

\clearpage

\subsubsection{Run Time and Token Usage}
All experiments are conducted on a server with an AMD EPYC 9654 
96-Core Processor and 14GB RAM. All methods are run single-threaded on CPU. 
Reported runtimes are approximate wall-clock times measured on Win95pts. Token usage is reported for LLM-augmented methods only, 
measured with \claude as the underlying LLM.
\begin{table}[ht]
\centering
\caption{Runtime and token usage comparison on Win95pts ($n{=}76$). 
Time is measured in wall-clock seconds. Token usage is reported for 
LLM-augmented methods only, measured using \claude. 
``---'' indicates no LLM queries are made.}
\label{tab:runtime}
\begin{tabular}{lrr}
\toprule
\textbf{Method} & \textbf{Time (s)} & \textbf{Token Usage} \\
\midrule
PC      &      4 & --- \\
FCI     &      4 & --- \\
\midrule
GES     &  164 & --- \\
GRaSP   & 33,454 & --- \\
CAMML & 4,077 & --- \\
\midrule
NOTEARS-MLP & 2,232 & --- \\
DAGMA & 16,655 & --- \\
\midrule
LLM-BFS & 637 & 2,971,861 \\
SCP & 25,116 & 3,029,612 \\
ET-MCMC & 224,485 & 12,895\\
LLM-MEC & 35,857 & 1,659 \\
LLM & 41 & 26,964 \\
Ours & 4387 & 26,267 \\
\bottomrule
\end{tabular}
\end{table}

\subsection{Details on Ablation Study}
We detail the implementation of each variant strategy in our ablation study.
\paragraph{Consensus Only.}                                                                                               Disputed edges are unconditionally discarded. The predicted graph contains exactly the edges on which all algorithms agree.
\paragraph{LLM Only.}                                                                                                     The LLM is queried independently over every variable with no consensus filtering and no algorithmic guidance. Domain 
context and variable interpretations are inferred in the same way as 
\emph{Ours}. For each variable $X_i$ ($i = 1, \ldots, n$), a single 
prompt asks \emph{``which variables does $X_i$ directly cause?''} and 
the LLM returns a list of effect variables. The predicted graph is the 
union of all returned out-edges across the $n$ calls.                                                                                                                
\paragraph{Majority Vote.}  All edges are resolved by majority vote: 
the option ($\text{fwd}$, $\text{rev}$, or $\text{none}$) receiving 
the most votes from the three algorithms is accepted, with ties broken 
in favor of $\text{none}$.

\paragraph{Cons + LLM.}
The consensus edges are kept as-is. The LLM 
is queried on disputed edges using the same bidirectional protocol as 
\emph{Ours} (Section~\ref{sec:calib}), producing identical LLM 
outputs. 
The LLM's answer is accepted directly without trust-weighted 
combination with algorithm votes.

\subsection{Additional Results on Model Selection Sensitivity}
\label{app:robust}
\begin{figure}[t]
  \centering
  \includegraphics[width=1\textwidth]{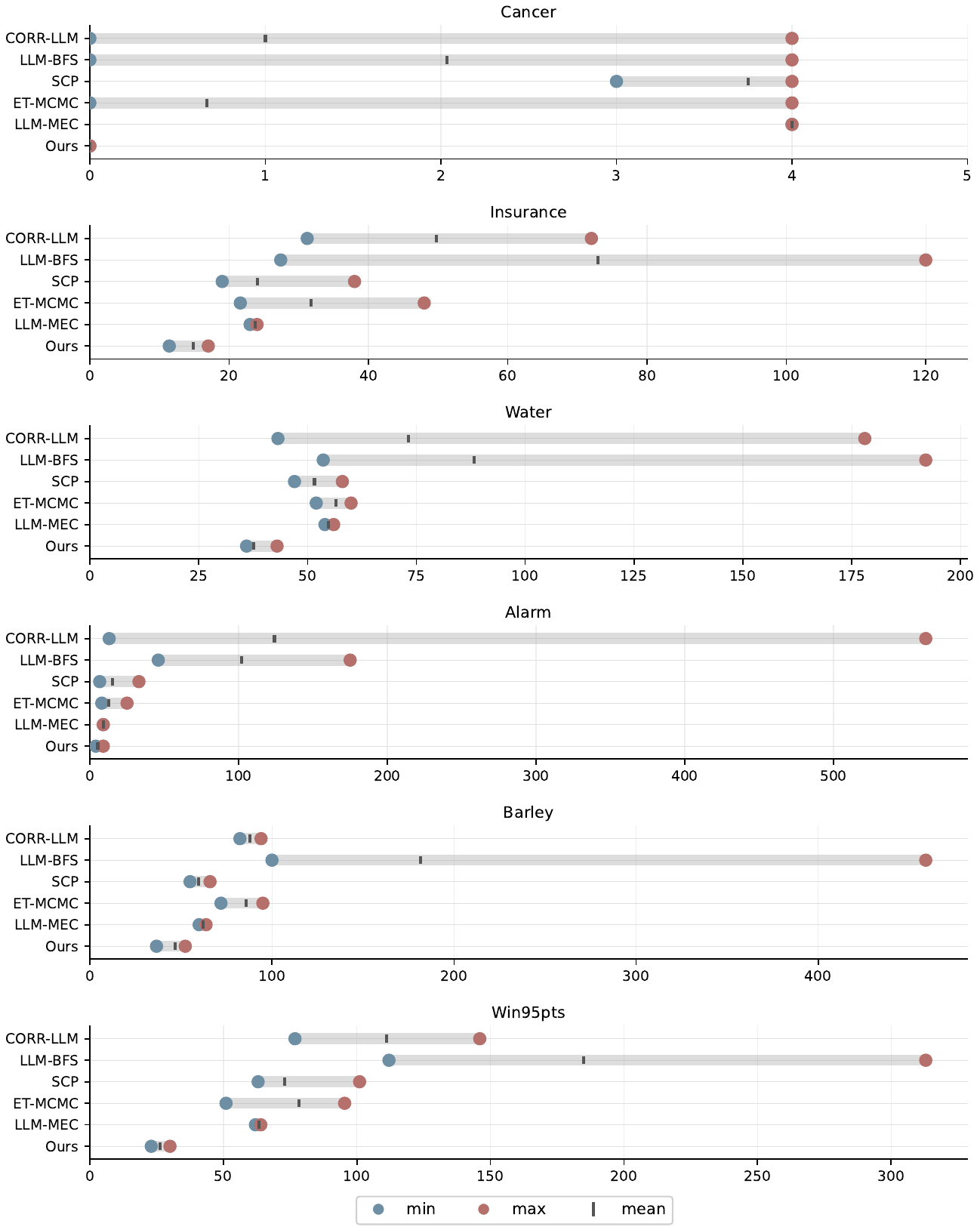}
  \caption{Cross-model robustness of LLM-augmented methods across 
datasets (SHD). Each horizontal bar spans the minimum to maximum SHD 
achieved across six LLMs, where shorter bars indicate greater 
robustness to backend selection.}
  \label{fig:app-robust-shd}
\end{figure}

\begin{figure}[t]
  \centering
  \includegraphics[width=1\textwidth]{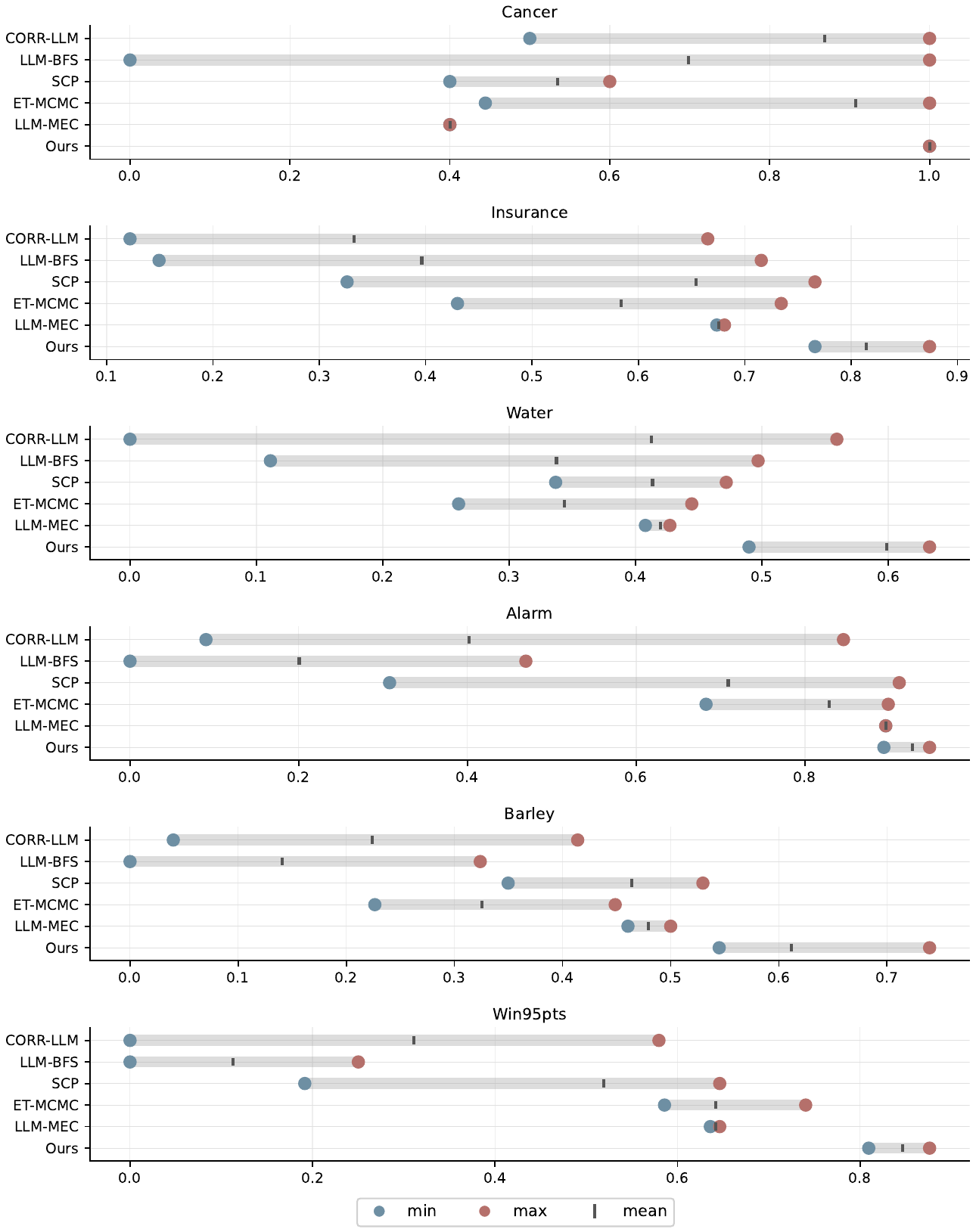}
  \caption{Cross-model robustness of LLM-augmented methods across 
datasets (F1). Each horizontal bar spans the minimum to maximum F1 
achieved across six LLMs.}
  \label{fig:app-robust-f1}
\end{figure}

\begin{figure}[t]
  \centering
  \includegraphics[width=1\textwidth]{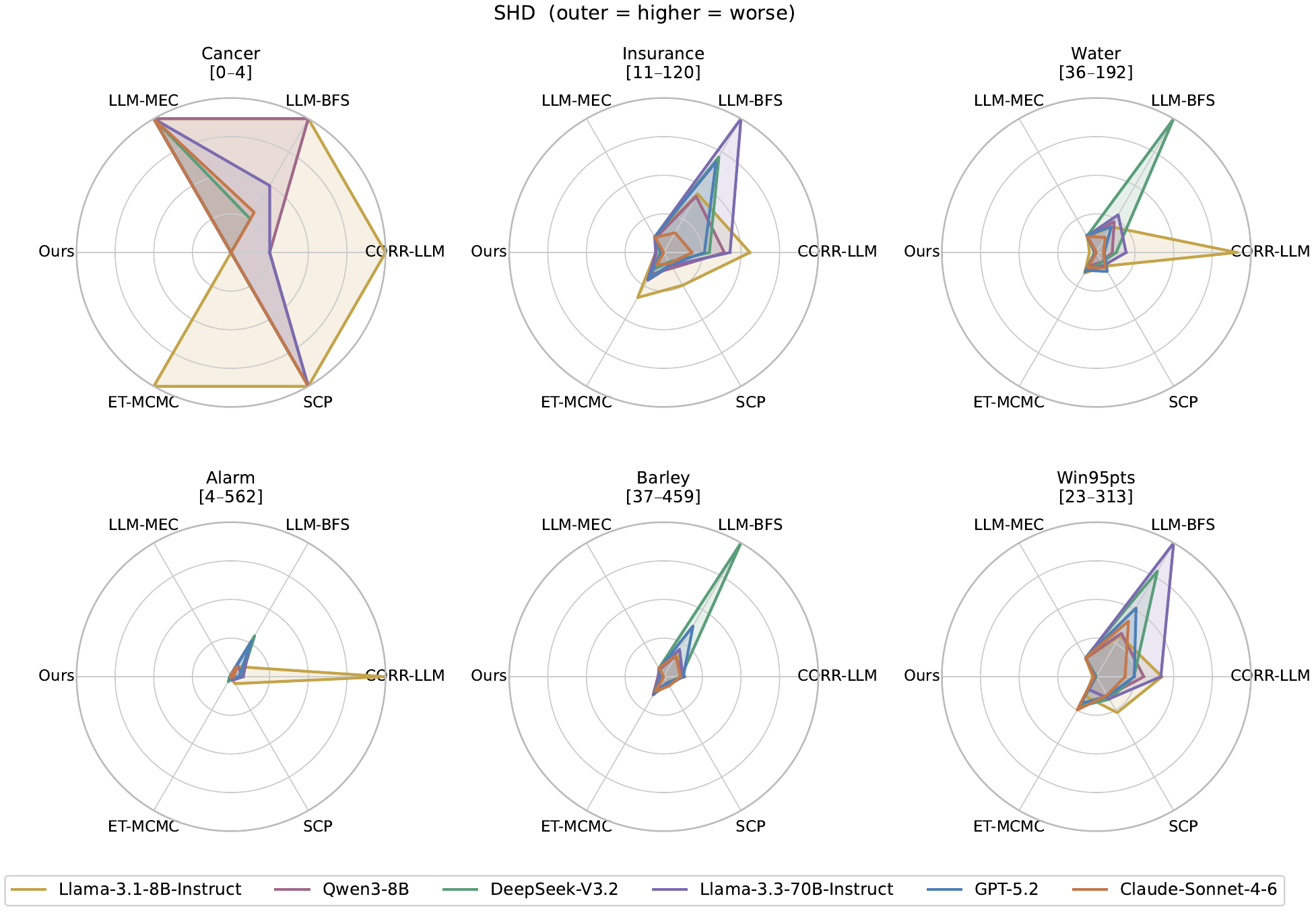}
  \caption{Cross-model radar of LLM-augmented methods across datasets (SHD). For each dataset sub-figure, the axis is scaled to the observed range across all models and methods for that dataset.}
  \label{fig:app-radar-shd}
\end{figure}

\begin{figure}[t]
  \centering
  \includegraphics[width=1\textwidth]{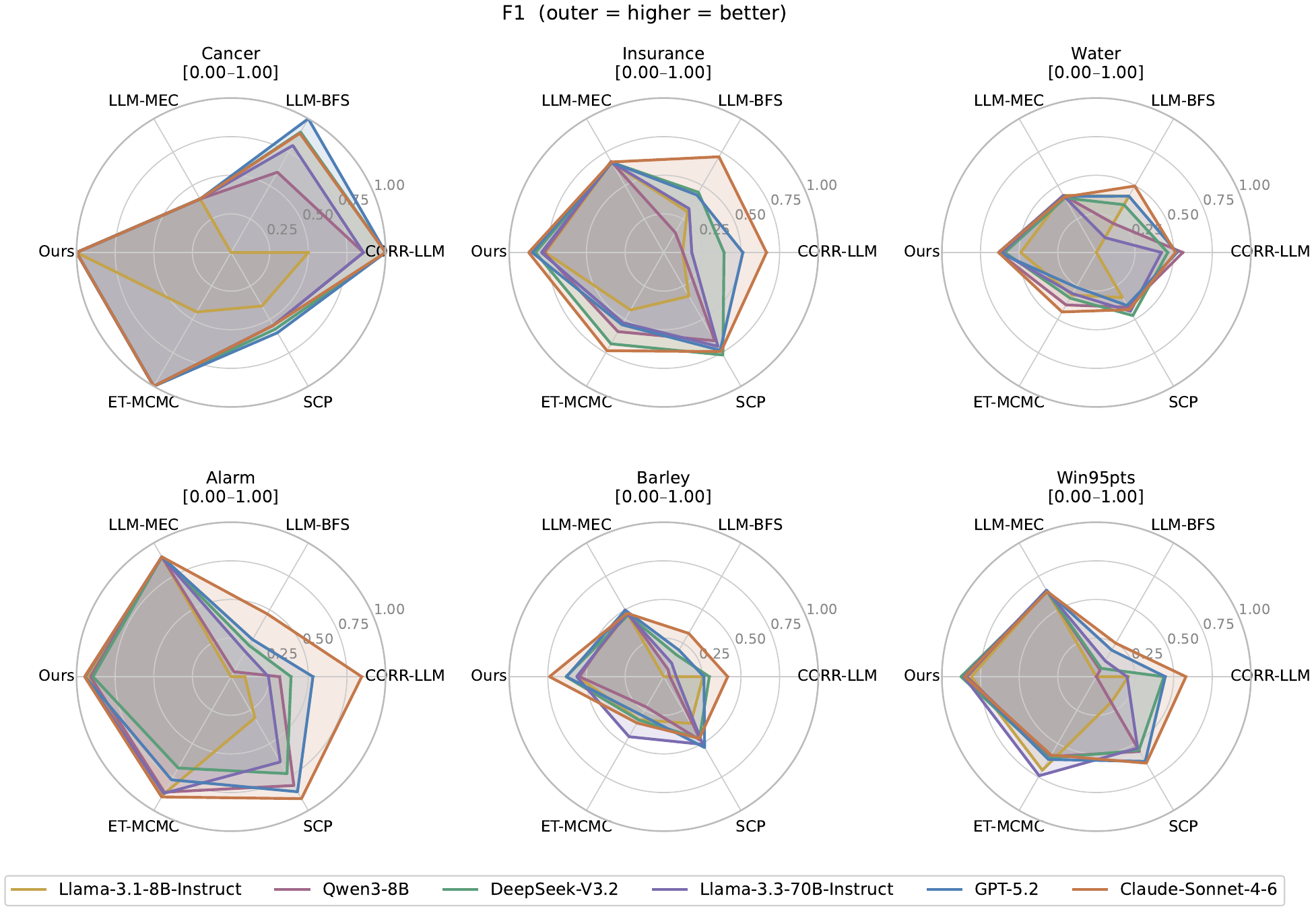}
  \caption{Cross-model radar of LLM-augmented methods across datasets (F1).}
  \label{fig:app-radar-f1}
\end{figure}

We provide more detailed cross-model robustness results in 
Figures~\ref{fig:app-robust-shd}--\ref{fig:app-robust-f1} and 
radar plots in Figures~\ref{fig:app-radar-shd}--\ref{fig:app-radar-f1}. 
Figures~\ref{fig:app-robust-shd} and~\ref{fig:app-robust-f1} show the 
SHD and F1 ranges across six LLMs for each method and dataset, 
where each dataset uses an independently scaled x-axis to make 
cross-model variance more distinguishable. 
Figures~\ref{fig:app-radar-shd} and~\ref{fig:app-radar-f1} show radar 
plots summarizing the performance of all LLMs across datasets 
and methods on SHD and F1 respectively.

\clearpage
\subsection{Consensus Filtering Rate and Accuracy}
Table~\ref{tab:consensus_filtering} reports the consensus filtering statistics across all six datasets. On larger graphs, consensus filtering resolves over 90\% of variable pairs with near-perfect accuracy, confirming the suitability of consensus edges as proxy ground truth for downstream calibration. The lower filtering rate on Cancer (n=5) is expected given its small number of variable pairs.

\begin{table}[h]                                                                                                                                                                 
  \centering                                                                                                                                                                       
  \caption{Consensus filtering statistics across datasets. Filter rate denotes the fraction of variable pairs resolved by consensus. Cons.\ Acc.\ denotes the fraction of consensus-resolved edges correctly classified.}                                                                                                                                                          
  \label{tab:consensus_filtering}                                                                                                                                                  
  \begin{tabular}{lrrrrrr}                                                                                                                                                         
  \toprule                                                                                                                                                                         
  Dataset   & Nodes & Total Pairs & Disputed & Filtered & Filter Rate & Cons.\ Acc. \\                                                                                             
  \midrule                                                                                                                                                                         
  Cancer    & 5  & 10    & 7  & 3     & 30.0\% & 100.0\% \\                                                                                                                        
  Insurance & 27 & 351   & 26 & 325   & 92.6\% & 97.8\%  \\                                                                                                                        
  Alarm     & 37 & 666   & 23 & 643   & 96.5\% & 99.8\%  \\                                                                                                                        
  Water     & 32 & 496   & 31 & 465   & 93.8\% & 92.7\%  \\                                                                                                                        
  Barley    & 48 & 1{,}128 & 70 & 1{,}058 & 93.8\% & 98.2\% \\
  Win95pts  & 76 & 2{,}850 & 96 & 2{,}754 & 96.6\% & 99.6\% \\                                                                                                                     
  \bottomrule                                                                                                                                                                      
  \end{tabular}                                                                                                                                                                    
  \end{table}  

\subsection{Evaluation on a Newly Constructed Graph}
\label{app:airport}
\begin{figure}[htbp]
  \centering
  \includegraphics[width=\linewidth]{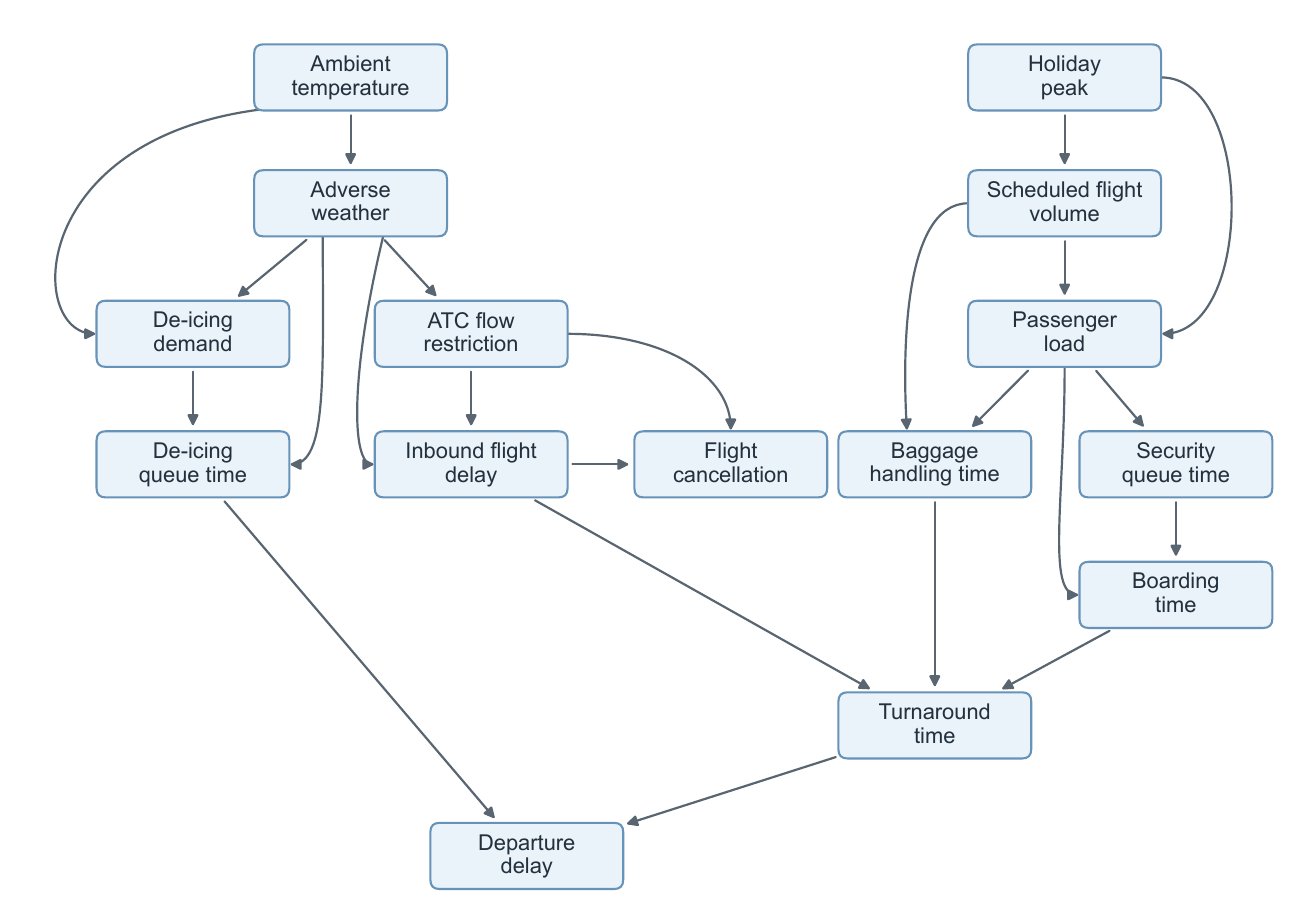}
  \caption{Ground-truth causal graph of Airport. Arrows indicate direct causal relationships. ATC denotes air traffic control.}
  \label{fig:airport-network}
\end{figure}

Beyond published benchmarks, we independently constructed \emph{Airport}, a synthetic Bayesian network with 15 variables and 23 directed edges.
Figure~\ref{fig:airport-network} shows its ground-truth causal structure.
The network combines three airport-related processes, with ambient temperature and holiday peaks as exogenous drivers. The weather branch links adverse weather, de-icing demand, and de-icing queue time. The air-traffic-control branch links flow restrictions, inbound flight delay, and flight cancellation. The terminal and ramp branch links scheduled flight volume to passenger load, which affects security queues, baggage handling, and boarding. Security queues affect boarding time; baggage handling and boarding both affect aircraft turnaround time. Turnaround time and de-icing queues, in turn, affect departure delay.

Each variable has three states (low, average, and high). The construction procedure fits discrete conditional probability tables to a discretized linear-Gaussian structural model. Table~\ref{tab:airport_results} reports \sys results on this sampled dataset. \sys achieves the best SHD and F1 and the tied-best SID on this newly constructed graph, indicating its robustness on non-public benchmarks.

\begin{table}[tp]
\centering
\caption{Results on \textbf{Airport} ($n=15$, $|E|=23$). F1 is reported in percent. SID is reported as an exact value for DAG outputs or as $[L,U]$ for distinct lower and upper bounds.}
\label{tab:airport_results}
{\small
\setlength{\tabcolsep}{10pt}
\begin{tabular}{lrrr}
\toprule
\textbf{Method} & \textbf{SHD} $\downarrow$ & \textbf{F1 (\%)} $\uparrow$ & \textbf{SID} $\downarrow$ \\
\midrule
PC            & 21 & 67.7 & $[8, 89]$ \\
FCI           & 22 & 66.7 & $[8, 102]$ \\
\midrule
GES           & 13 & 75.0 & $[17, 70]$ \\
GRaSP         & 20 & 59.6 & $[50, 93]$ \\
CAMML         & 5  & 78.3 & 38 \\
\midrule
NOTEARS-MLP   & 8  & 68.1 & 51 \\
DAGMA         & 6  & 77.3 & 54 \\
\midrule
CORR-LLM      & 5  & 89.4 & 8 \\
LLM-BFS       & 11 & 78.4 & 15 \\
SCP           & 7  & 86.3 & $[8, 44]$ \\
ET-MCMC       & 5  & 78.3 & 37 \\
LLM-MEC       & 16 & 73.3 & $[8, 75]$ \\
\midrule
Ours          & \textbf{2} & \textbf{95.7} & 8 \\
\bottomrule
\end{tabular}
}
\end{table}

\end{document}

%% file: math_commands.tex
\usepackage{amsmath,amsfonts,bm}

\def\eqref#1{equation~\ref{#1}}

\def\1{\bm{1}}

\DeclareMathAlphabet{\mathsfit}{\encodingdefault}{\sfdefault}{m}{sl}
\SetMathAlphabet{\mathsfit}{bold}{\encodingdefault}{\sfdefault}{bx}{n}



%% file: iclr2027_conference.bib
@inproceedings{abdulaal2023causal,
  title={Causal modelling agents: Causal graph discovery through synergising metadata-and data-driven reasoning},
  author={Abdulaal, Ahmed and Montana-Brown, Nina and He, Tiantian and Ijishakin, Ayodeji and Drobnjak, Ivana and Castro, Daniel C and Alexander, Daniel C and others},
  booktitle={The Twelfth International Conference on Learning Representations},
  year={2024}
}

@inproceedings{10.5555/2074158.2074215,
author = {Spirtes, Peter and Meek, Christopher and Richardson, Thomas},
title = {Causal inference in the presence of latent variables and selection bias},
year = {1995},
booktitle = {Proceedings of the Eleventh Conference on Uncertainty in Artificial Intelligence},
pages = {499–506},
numpages = {8}
}

@inproceedings{long2023causal,
  title     = {Causal Discovery with Language Models as Imperfect Experts},
  author    = {Long, Stephanie and Pich{\'e}, Alexandre and Zantedeschi, Valentina and Schuster, Tibor and Drouin, Alexandre},
  booktitle = {ICML 2023 Workshop on Structured Probabilistic Inference and Generative Modeling}, 
  year      = {2023}
}

@ARTICLE{10966043,
  author={Ban, Taiyu and Chen, Lyuzhou and Lyu, Derui and Wang, Xiangyu and Zhu, Qinrui and Tu, Qiang and Chen, Huanhuan},
  journal={IEEE Transactions on Artificial Intelligence}, 
  title={Integrating Large Language Model for Improved Causal Discovery}, 
  year={2025},
  volume={6},
  number={11},
  pages={3030-3042},
  doi={10.1109/TAI.2025.3560927}}

@inproceedings{roy2025causal,
  title={Causal-LLM: A Unified One-Shot Framework for Prompt-and Data-Driven Causal Graph Discovery},
  author={Roy, Amartya and Devharish, N and Ganguly, Shreya and Ghosh, Kripabandhu},
  booktitle={Findings of the Association for Computational Linguistics: EMNLP 2025},
  pages={8259--8279},
  year={2025}
}

@inproceedings{
jiralerspong2024efficient,
title={Efficient Causal Graph Discovery Using Large Language Models},
author={Thomas Jiralerspong and Xiaoyin Chen and Yash More and Vedant Shah and Yoshua Bengio},
booktitle={ICLR 2024 Workshop: How Far Are We From AGI},
year={2024},
url={https://openreview.net/forum?id=5RBUTx75yr}
}

@inproceedings{
vashishtha2025causal,
title={Causal Order: The Key to Leveraging Imperfect Experts in Causal Inference},
author={Aniket Vashishtha and Abbavaram Gowtham Reddy and Abhinav Kumar and Saketh Bachu and Vineeth N. Balasubramanian and Amit Sharma},
booktitle={The Thirteenth International Conference on Learning Representations},
year={2025},
url={https://openreview.net/forum?id=9juyeCqL0u}
}

@article{
takayama2025integrating,
title={Integrating Large Language Models in Causal Discovery: A Statistical Causal Approach},
author={Masayuki Takayama and Tadahisa Okuda and Thong Pham and Tatsuyoshi Ikenoue and Shingo Fukuma and Shohei Shimizu and Akiyoshi Sannai},
journal={Transactions on Machine Learning Research},
issn={2835-8856},
year={2025},
url={https://openreview.net/forum?id=Reh1S8rxfh},
note={}
}

@article{scutari2010learning,
  title={Learning Bayesian networks with the bnlearn R package},
  author={Scutari, Marco},
  journal={Journal of statistical software},
  volume={35},
  pages={1--22},
  year={2010}
}

@inproceedings{zheng2018dags,
    author = {Zheng, Xun and Aragam, Bryon and Ravikumar, Pradeep and Xing, Eric P.},
    booktitle = {Advances in Neural Information Processing Systems},
    title = {{DAGs with NO TEARS: Continuous Optimization for Structure Learning}},
    year = {2018}
}

@book{spirtes2000causation,
  title={Causation, Prediction, and Search},
  author={Spirtes, Peter and Glymour, Clark and Scheines, Richard},
  year={2000},
  edition={2nd},
  publisher={The MIT Press},
  address={Cambridge, MA}
}

@book{pearl2009causality,
  title={Causality},
  author={Pearl, Judea},
  year={2009},
  publisher={Cambridge University Press}
}

@article{peng2026causcientist,
  title={CauScientist: Teaching LLMs to Respect Data for Causal Discovery},
  author={Peng, Bo and Chen, Sirui and Xu, Lei and Lu, Chaochao},
  journal={arXiv preprint arXiv:2601.13614},
  year={2026}
}

@inproceedings{Wallace1996CausalDV,
  title={Causal Discovery via MML},
  author={Wallace, Chris S and Korb, Kevin B and Dai, Honghua},
  booktitle={Proceedings of the Thirteenth International Conference on Machine Learning},
  pages={516--524},
  year={1996}
}

@inproceedings{bello2022dagma,
  title={{DAGMA: Learning DAGs via M-matrices and a Log-Determinant Acyclicity Characterization}},
  author={Bello, Kevin and Aragam, Bryon and Ravikumar, Pradeep},
  booktitle={Advances in Neural Information Processing Systems},
  volume={35},
  pages={8226--8239},
  year={2022}
}

@inproceedings{lam2022grasp,
  title={Greedy relaxations of the sparsest permutation algorithm},
  author={Lam, Wai-Yin and Andrews, Bryan and Ramsey, Joseph},
  booktitle={Uncertainty in Artificial Intelligence},
  pages={1052--1062},
  year={2022},
  organization={PMLR}
}

@inproceedings{andrews2023fast,
  title={Fast scalable and accurate discovery of dags using the best order score search and grow shrink trees},
  author={Andrews, Bryan and Ramsey, Joseph and Sanchez Romero, Ruben and Camchong, Jazmin and Kummerfeld, Erich},
  booktitle ={Advances in Neural Information Processing Systems},
  volume={36},
  pages={63945--63956},
  year={2023}
}

@article{chickering2002optimal,
  title   = {Optimal structure identification with greedy search},
  author  = {Chickering, David Maxwell},
  journal = {Journal of machine learning research},
  volume  = {3},
  pages   = {507--554},
  year    = {2002}
}

@inproceedings{zheng2020learning,
  title        = {Learning sparse nonparametric DAGs},
  author       = {Zheng, Xun and Dan, Chen and Aragam, Bryon and Ravikumar, Pradeep and Xing, Eric},
  booktitle    = {International Conference on Artificial Intelligence and Statistics},
  pages        = {3414--3425},
  year         = {2020},
  organization = {PMLR}
}

@article{peters2015structural,
  title     = {Structural intervention distance for evaluating causal graphs},
  author    = {Peters, Jonas and B{\"u}hlmann, Peter},
  journal   = {Neural computation},
  volume    = {27},
  number    = {3},
  pages     = {771--799},
  year      = {2015},
  publisher = {MIT Press}
}

@book{korb2010bayesian,
  title={Bayesian Artificial Intelligence},
  author={Korb, Kevin B and Nicholson, Ann E},
  edition={2nd},
  year={2010},
  publisher={CRC Press}
}

@article{binder1997adaptive,
  title={Adaptive probabilistic networks with hidden variables},
  author={Binder, John and Koller, Daphne and Russell, Stuart and Kanazawa, Keiji},
  journal={Machine Learning},
  volume={29},
  number={2},
  pages={213--244},
  year={1997},
  publisher={Springer}
}

@techreport{jensen1989expert,
  title={An expert system for control of waste water treatment—a pilot project},
  author={Jensen, FV and Kj{\ae}rulff, U and Olesen, KG and Pedersen, J},
  year={1989},
  institution={Technical report, Judex Datasystemer A/S, Aalborg, 1989. In Danish}
}

@inproceedings{beinlich1989alarm,
  title={The ALARM monitoring system: A case study with two probabilistic inference techniques for belief networks},
  author={Beinlich, Ingo A and Suermondt, Henri Jacques and Chavez, R Martin and Cooper, Gregory F},
  booktitle={AIME 89: Second European Conference on Artificial Intelligence in Medicine, London, August 29th--31st 1989. Proceedings},
  pages={247--256},
  year={1989},
  organization={Springer}
}

@article{kristensen2002use,
  title     = {The use of a Bayesian network in the design of a decision support system for growing malting barley without use of pesticides},
  author    = {Kristensen, Kristian and Rasmussen, Ilse A},
  journal   = {Computers and Electronics in Agriculture},
  volume    = {33},
  number    = {3},
  pages     = {197--217},
  year      = {2002},
  publisher = {Elsevier}
}

@article{heckerman1995decision,
  title={Decision-theoretic troubleshooting},
  author={Heckerman, David and Breese, John S and Rommelse, Koos},
  journal={Communications of the ACM},
  volume={38},
  number={3},
  pages={49--57},
  year={1995},
  publisher={ACM New York, NY, USA}
}

@inproceedings{xiong2024can,
  title={Can llms express their uncertainty? an empirical evaluation of confidence elicitation in llms},
  author={Xiong, Miao and Hu, Zhiyuan and Lu, Xinyang and Li, Yifei and Fu, Jie and He, Junxian and Hooi, Bryan},
  booktitle={International Conference on Learning Representations},
  volume={2024},
  pages={23650--23678},
  year={2024}
}

@article{pearl2018book,
  title={The book of why},
  author={Pearl, Judea and Mackenzie, Dana},
  year={2018},
  publisher={Allen Lane London}
}

@article{yao2021survey,
  title     = {A survey on causal inference},
  author    = {Yao, Liuyi and Chu, Zhixuan and Li, Sheng and Li, Yaliang and Gao, Jing and Zhang, Aidong},
  journal   = {ACM Transactions on Knowledge Discovery from Data (TKDD)},
  volume    = {15},
  number    = {5},
  pages     = {1--46},
  year      = {2021},
  publisher = {ACM New York, NY, USA}
}

@article{ankan2024pgmpy,
  author  = {Ankur Ankan and Johannes Textor},
  title   = {pgmpy: A Python Toolkit for Bayesian Networks},
  journal = {Journal of Machine Learning Research},
  year    = {2024},
  volume  = {25},
  number  = {265},
  pages   = {1--8},
  url     = {http://jmlr.org/papers/v25/23-0487.html}
}

@article{scikit-learn,
  title   = {Scikit-learn: Machine Learning in {P}ython},
  author  = {Pedregosa, F. and Varoquaux, G. and Gramfort, A. and Michel, V.
             and Thirion, B. and Grisel, O. and Blondel, M. and Prettenhofer, P.
             and Weiss, R. and Dubourg, V. and Vanderplas, J. and Passos, A. and
             Cournapeau, D. and Brucher, M. and Perrot, M. and Duchesnay, E.},
  journal = {Journal of Machine Learning Research},
  volume  = {12},
  pages   = {2825--2830},
  year    = {2011}
}

@article{zheng2024causal,
  title   = {Causal-learn: Causal discovery in python},
  author  = {Zheng, Yujia and Huang, Biwei and Chen, Wei and Ramsey, Joseph and Gong, Mingming and Cai, Ruichu and Shimizu, Shohei and Spirtes, Peter and Zhang, Kun},
  journal = {Journal of Machine Learning Research},
  volume  = {25},
  number  = {60},
  pages   = {1--8},
  year    = {2024}
}

@article{zhang2021gcastle,
  title={gcastle: A python toolbox for causal discovery},
  author={Zhang, Keli and Zhu, Shengyu and Kalander, Marcus and Ng, Ignavier and Ye, Junjian and Chen, Zhitang and Pan, Lujia},
  journal={arXiv preprint arXiv:2111.15155},
  year={2021}
}

@article{
wu2025sample,
title={Sample, estimate, aggregate: A recipe for causal discovery foundation models},
author={Menghua Wu and Yujia Bao and Regina Barzilay and Tommi Jaakkola},
journal={Transactions on Machine Learning Research},
issn={2835-8856},
year={2025},
url={https://openreview.net/forum?id=h434zx5SX0},
note={}
}

@inproceedings{peng2026causcale,
  title={CauScale: Neural Causal Discovery at Scale},
  author={Peng, Bo and Chen, Sirui and Tian, Jiaguo and Qiao, Yu and Lu, Chaochao},
  booktitle={Proceedings of the 43rd International Conference on Machine Learning},
  year={2026}
}

@article{pawlowski2020deep,
  title={Deep structural causal models for tractable counterfactual inference},
  author={Pawlowski, Nick and Coelho de Castro, Daniel and Glocker, Ben},
  journal={Advances in neural information processing systems},
  volume={33},
  pages={857--869},
  year={2020}
}

@incollection{verma1990equivalence,
  title={Equivalence and synthesis of causal models},
  author={Verma, Thomas S and Pearl, Judea},
  booktitle={Probabilistic and causal inference: The works of Judea Pearl},
  pages={221--236},
  year={2022}
}
